\documentclass[10pt,journal,compsoc]{IEEEtran}

\ifCLASSOPTIONcompsoc
  \usepackage[nocompress]{cite}
\else
  \usepackage{cite}
\fi
\ifCLASSINFOpdf
  \usepackage[pdftex]{graphicx}
  \usepackage{tikz}
  \usetikzlibrary{arrows.meta}
  \usepackage{pgfplots}
  \graphicspath{{illustrations/}}
  \usepgfplotslibrary{fillbetween}
  \definecolor{dlrred}{RGB}{179,63,61}
  \definecolor{dlrblue}{RGB}{0,117,187}
  \definecolor{dlryellow}{RGB}{234,184,24}
\else
\fi
\usepackage{amsmath}
\usepackage{amssymb}
\usepackage{mathtools}
\usepackage{units}
\DeclareMathOperator{\Ad}{Ad}

\usepackage[nolist]{acronym}
\renewcommand{\IEEEiedlistdecl}{\IEEEsetlabelwidth{SONET}}
\begin{acronym}
	\acro{ICP}[ICP]{Iterative Closest Point}
 	\acro{PDF}[PDF]{probability density function}
 	\acro{6DoF}[6DoF]{six degrees of freedom}
 	\acro{CNN}[CNN]{convolutional neural network}
 	\acro{EKF}[EKF]{Extended Kalman Filter}
 	\acro{UKF}[UKF]{Unscented Kalman Filter}
 	\acro{SGM}[SGM]{Semi-Global Matching}
 	\acro{RTB}[RTB]{Robot Tracking Benchmark}
 	\acro{ADD}[ADD]{average-distance}
 	\acro{ADD-S}[ADD-S]{average-shortest-distance}
\end{acronym}
\renewcommand{\IEEEiedlistdecl}{\relax}
\usepackage{tabularx}
\usepackage{booktabs}
\usepackage{multirow}

\ifCLASSOPTIONcompsoc
 \usepackage[caption=false,font=footnotesize,labelfont=sf,textfont=sf]{subfig}
\else
 \usepackage[caption=false,font=footnotesize]{subfig}
\fi
\usepackage{url}
\newcommand\MYhyperrefoptions{bookmarks=true,bookmarksnumbered=true,
pdfpagemode={UseOutlines},plainpages=false,pdfpagelabels=true,
colorlinks=true,linkcolor={black},citecolor={black},urlcolor={black},
pdftitle={A Multi-body Tracking Framework - From Rigid Objects to Kinematic Structures},
}
\ifCLASSINFOpdf
\usepackage[\MYhyperrefoptions,pdftex]{hyperref}
\else
\usepackage[\MYhyperrefoptions,breaklinks=true,dvips]{hyperref}
\usepackage{breakurl}
\fi
\begin{document}
%
\title{A Multi-body Tracking Framework - \\From Rigid Objects to Kinematic Structures}
%
%
%
%

\author{Manuel~Stoiber, 
        Martin~Sundermeyer, 
        Wout~Boerdijk, 
        and~Rudolph~Triebel
\IEEEcompsocitemizethanks{\IEEEcompsocthanksitem M.~Stoiber, M.~Sundermeyer, W.~Boerdijk, and R.~Triebel are with the Institute of Robotics and Mechatronics, German Aerospace Center (DLR), 82234 Wessling, Germany and with the Department of Informatics, Technical University of Munich (TUM), 80333 Munich, Germany. \protect\\
E-mail: \tt{\{firstname.lastname\}@dlr.de}}
}

%
%

\markboth{
}%
{Stoiber \MakeLowercase{\textit{et al.}}: A Multi-body Tracking Framework - From Rigid Objects to Kinematic Structures}
%



\IEEEtitleabstractindextext{%
\begin{abstract}
Kinematic structures are very common in the real world.
They range from simple articulated objects to complex mechanical systems.
However, despite their relevance, most model-based 3D tracking methods only consider rigid objects.
To overcome this limitation, we propose a flexible framework that allows the extension of existing \acs{6DoF} algorithms to kinematic structures.
Our approach focuses on methods that employ Newton-like optimization techniques, which are widely used in object tracking.
The framework considers both tree-like and closed kinematic structures and allows a flexible configuration of joints and constraints.
To project equations from individual rigid bodies to a multi-body system, Jacobians are used.
For closed kinematic chains, a novel formulation that features Lagrange multipliers is developed.
In a detailed mathematical proof, we show that our constraint formulation leads to an exact kinematic solution and converges in a single iteration.
Based on the proposed framework, we extend \textit{ICG}, which is a state-of-the-art rigid object tracking algorithm, to multi-body tracking.
For the evaluation, we create a highly-realistic synthetic dataset that features a large number of sequences and various robots.
Based on this dataset, we conduct a wide variety of experiments that demonstrate the excellent performance of the developed framework and our multi-body tracker.

\end{abstract}

}

\maketitle

\IEEEdisplaynontitleabstractindextext

%
\IEEEpeerreviewmaketitle

\ifCLASSOPTIONcompsoc
\IEEEraisesectionheading{\section{Introduction}\label{sec:in}}
\else
\section{Introduction}
\label{sec:in}
\fi


\IEEEPARstart{T}{racking} objects in 3D space and estimating their \ac{6DoF} pose is an essential task in computer vision.
The goal is to provide the position and orientation with respect to the camera at high frequency, given consecutive images and the 3D model.
For kinematic structures, the object's configuration is also required.
It directly corresponds to the \ac{6DoF} pose of all individual rigid bodies in the system.
Typical applications of 3D object tracking range from augmented reality to robotics.
For augmented reality, the pose and configuration are used to superimpose digital information on real-world objects.
In robotics, continuous estimates are needed to facilitate complex manipulation tasks and allow robots to react to unpredicted changes in the environment.
In many such applications, objects are not rigid.
Instead, they consist of multiple connected bodies that form kinematic structures.
Typical examples range from various tools, furniture, and appliances to machinery and robots.

For 3D object tracking, multiple challenges exist, including occlusions, cluttered environments, motion blur, textureless surfaces, object symmetries, computational complexity, and many more.
As a consequence, a wide variety of approaches have been developed \cite{Lepetit2005, Yilmaz2006}.
Methods are typically differentiated by their use of keypoints \cite{Vacchetti2004, Rublee2011}, explicit edges \cite{Harris1990, Drummond2002b}, direct optimization \cite{Lucas1981, Crivellaro2014}, deep learning \cite{Wen2020, Deng2021}, image regions \cite{Prisacariu2012, Stoiber2021}, and depth information \cite{Schmidt2015, Issac2016}.
Depending on the application, all techniques have their own advantages and disadvantages, showing different properties with respect to accuracy, robustness, and runtime.
However, most state-of-the-art methods only implement the tracking of rigid objects without considering multi-body systems.
While, in theory, such algorithms could track bodies independently, in most cases, the task is unfeasible without kinematic information.
As a consequence, many methods with desirable properties can not be used in applications that feature kinematic structures.

\begin{figure}[t]
	\centering
	\includegraphics[width=\linewidth]{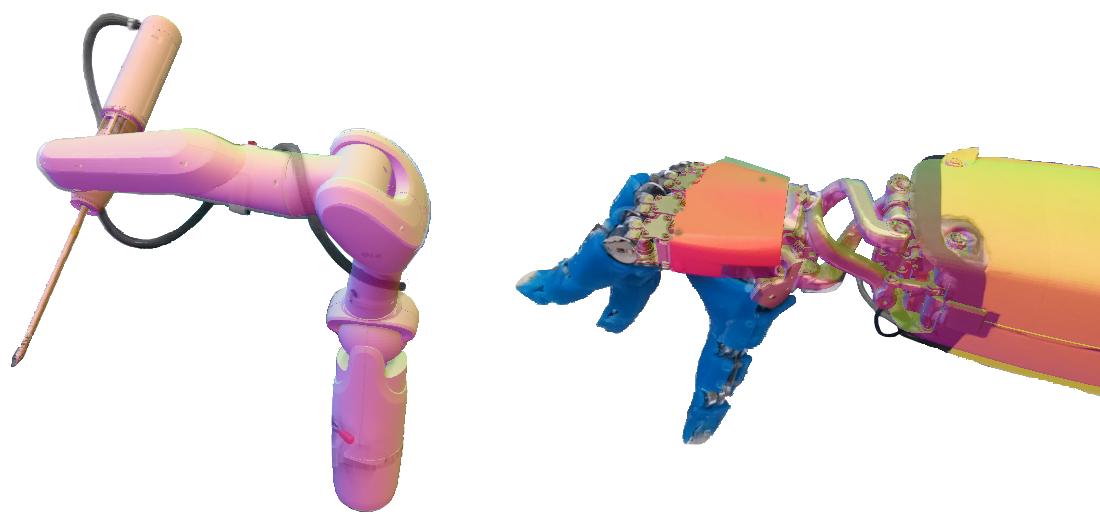}
	\caption{
		Real-world examples that show the tracking of kinematic structures.
		To illustrate the predicted pose and configuration, the object model is rendered as an overlay.
		The image on the left shows the tracking of the medical robot \textit{MIRO}, which consists of a 7 degrees-of-freedom chain structure.
		In the right image, the palm, wrist, and forearm of the humanoid robot \textit{David} are tracked.
		The kinematic structure includes multiple closed chains and consists of 12 individual bodies.
	}\label{fig:in00}
\end{figure}

To overcome this limitation, we propose a flexible framework that allows the extension of rigid object tracking methods to kinematic structures.
Our approach focuses on Newton-like optimization, which is widely used in various tracking techniques.
The resulting framework allows a flexible configuration of multi-body systems with various joints and constraints.
To the best of our knowledge, it is the first that is able to accurately model closed kinematic chains. 
Based on the developed framework, we extend our previous work \textit{ICG} \cite{Stoiber2022} to multi-body tracking.
\textit{ICG} combines region and depth information, is highly efficient, and performs well for a wide variety of objects and scenarios.
Example images of the extended method are shown in Fig.\,\ref{fig:in00}.
In a thorough evaluation on a novel, highly-realistic synthetic dataset, we demonstrate the quality and consistency of the developed approach.
In summary, our contributions are as follows:
\begin{itemize}
	\item A unified framework that combines the projection to a minimal parameterization and the application of pose constraints in a single formulation.
	\item Constraint equations that consider pose differences and that directly operate on the rotation vector, covering the entire space of possible rotations.
	\item Thorough mathematical derivations with a proof that shows that constraints enforce an exact solution for which pose differences converge in a single iteration.
	\item A multi-body 3D tracking method that extends \textit{ICG} and considers both depth and region information.
	\item A realistic synthetic dataset that features a large number of sequences with various moving robots.
\end{itemize}

In the following, we first discuss related work on multi-body object tracking.
Subsequently, our multi-body tracking framework is introduced, which allows to extend tracking algorithms that use Newton-like optimization techniques.
The framework is formulated without considering a particular parameterization.
We then focus on the popular axis-angle representation and derive all equations required for joints and constraints.
In addition, properties of our constraint formulation are analyzed mathematically.
This is followed by a description of all modifications to \textit{ICG} that are required for multi-body tracking.
Finally, after introducing the created dataset, a thorough evaluation of the developed framework and multi-body tracker is presented.
Note that both the source code of our tracker and the developed dataset are publicly available.\footnote{\label{foo:dataset}\url{https://zenodo.org/record/7548537}}\textsuperscript{\hspace{-0.1em},}\footnote{\label{foo:code}\url{https://github.com/DLR-RM/3DObjectTracking}}


\section{Related Work}\label{sec:rw}
In the following section, we provide a detailed overview of trackers for multi-body objects with a focus on gradient-based techniques and rigid bodies.
Algorithms for elastic objects, such as human hands or bodies, are only briefly discussed.
While some early methods \cite{Hogg1983, Gavrila1996} employed search algorithms to determine the pose and configuration of objects, Lowe \cite{Lowe1991} proposed the use of nonlinear least-squares optimization.
He showed that the approach is both robust and efficient, and allows to determine the state of articulated objects for large frame-to-frame motion.
Similar techniques were later adopted for model-based hand tracking \cite{Rehg1994} and to recover human body configurations \cite{Bregler1998}.
Also, \cite{Nickels2001} proposed the tracking of articulated objects using feature points in an \textit{\ac{EKF}}.
Finally, a more formal derivation motivated by the Lagrange-d'Alembert formulation in classical physics was also presented \cite{Comport2007}.
In all those methods, Jacobian matrices that describe the variation of individual points or measurements for a minimal set of parameters are derived.

In contrast to the use of a minimal parameterization, Drummond and Cipolla \cite{Drummond2002b} proposed an approach that adopts Lagrange multipliers in combination with a Lie group formalism.
Their method first predicts the \ac{6DoF} pose variation of all rigid bodies independently and later ensures compatibility using velocity constraints.
A detailed comparison of this \textit{postimposed} approach compared to \textit{preimposed} methods using Jacobians was later presented \cite{Campos2006}.
The experiments demonstrate that results for the two techniques are identical.
Also, they showed that while \textit{postimposed} constraints are very efficient for simple kinematic chains and allow temporary constraints, \textit{preimposed} methods scale better for more complex tree-like kinematic structures.
In contrast to those mathematically exact constraints, approaches that minimize the Mahalanobis distance \cite{Demirdjian2003} or that introduce soft revolute joints \cite{Mundermann2007} were also proposed.

Based on the presented principles to model kinematic structures, various methods that employ different tracking techniques have been developed.
Approaches that consider depth information using the \textit{\ac{ICP}} algorithm \cite{Besl1992} were proposed by \cite{Dewaele2004} and \cite{Pellegrini2008}.
Later, \cite{Brox2010} fused information from region fitting, dense optical flow, and \textit{SIFT} features \cite{Lowe2004}.
Similarly, \cite{Krainin2011} combined \textit{SIFT} features and dense color information with \textit{\ac{ICP}}-like depth measurements.
Information from depth sensors and joint encoders was utilized by \cite{Klingensmith2013}.
Also, joint positions that are predicted using pixel-wise part classification on depth images were used \cite{Bohg2014}.
Similarly, \cite{Rauch2018} employed pixel-wise part classification but directly considered depth values instead of joint positions.
Lately, an algorithm that fuses region and dense color information was also presented \cite{Liu2021a}.
With \textit{DART} \cite{Schmidt2015, Schmidt2015a}, a general user-friendly method was developed that extends \ac{6DoF} tracking with signed distance functions \cite{Fitzgibbon2003} to articulated objects.
Later, the integration of \textit{DART} with a tactile sensor was studied by \cite{Izatt2017}.
A similarly general method that uses dense optical flow and \textit{\ac{ICP}}-like depth measurements was developed by \cite{Pauwels2014b}.
It was later renamed to \textit{SimTrack} \cite{Pauwels2015}.
Except for \textit{SimTrack}, which implements the \textit{postimposed} constraints of Drummond and Cipolla \cite{Drummond2002b}, most methods use \textit{preimposed} formulations similar to the approach of Lowe \cite{Lowe1991}.

Apart from gradient-based methods, other optimization techniques were also adopted for the tracking of multi-body objects. 
In the work of \cite{Hebert2012}, an \textit{\ac{UKF}} was used to fuse different sources of information.
A method that employs a \textit{Random Forest} to directly regress to joint angles was proposed by \cite{Widmaier2016}.
Also, \cite{Cifuentes2017} used the \textit{Coordinate Particle Filter} \cite{Wuethrich2015} to track robot manipulators, considering information from both depth images and joint measurements.
Finally, deep learning was employed to recognize 2D keypoints on a robot \cite{Zuo2019} or use a render-and-compare strategy for pose estimation \cite{Labbe2021}.
In addition to methods for rigid objects, a large amount of work on the tracking of human hands and bodies exists.
While deep learning has become highly popular for those tasks \cite{Zimmermann2017,Kanazawa2018, Ge2019}, some model-based methods still use the original formulation of Lowe to consider the underlying kinematic structure \cite{Tan2016, Taylor2016, Han2020}.
However, because techniques like keypoint detection and regularization are highly optimized for their respective domains, there is no straightforward application of such algorithms to arbitrary multi-body systems.


\section{Multi-body Tracking}\label{sec:m}
In the following, we derive a flexible framework that allows to extend tracking algorithms from rigid objects to kinematic structures.
For this, the general architecture of methods that are compatible with our approach is discussed.
This is followed by an extension to tree-like structures and, finally, to closed kinematic structures.
The provided formulation is very general and allows different parameterizations of poses, joints, and constraints.

\subsection{Rigid Objects}\label{ssec:m1}
For the task of 3D object tracking, a wide variety of algorithms that use different sources of information have been developed.
Based on the considered data, most methods derive an energy function $E(\pmb{\theta})$ or \ac{PDF} $p(\pmb{\theta})$ that depends on the pose variation vector $\pmb{\theta}$.
Given such a function, the pose that best explains the considered data is found by minimizing the energy or maximizing the probability.
While different techniques have been employed for this optimization task, the following framework focuses on Newton-like methods.

In general, Newton-like optimization iteratively estimates the variation vector and updates the object pose.
The estimated pose variation $\pmb{\hat{\theta}}$ is thereby calculated as follows
\begin{equation} \label{eq:m10}
	\pmb{\hat{\theta}} = - \pmb{H}^{-1}\pmb{g},
\end{equation}
where the gradient vector $\pmb{g}$ and the Hessian matrix $\pmb{H}$ are the first- and second-order derivatives of the energy function $E(\pmb{\theta})$ or the negative logarithmic probability $-\ln(p(\pmb{\theta}))$.
In the following, we derive our approach for the energy function $E(\pmb{\theta})$.
Note, however, that with the relation \mbox{$E(\pmb{\theta}) = -\ln(p(\pmb{\theta}))$}, exactly the same equations can be used for \acp{PDF}.
Finally, (\ref{eq:m10}) is not only limited to classical Newton optimization with an exact Hessian matrix but also allows the use of approximations from Gauss-Newton or quasi-Newton methods.
As a consequence, many popular algorithms, such as \textit{RAPID} \cite{Harris1990},  \textit{PWP3D} \cite{Prisacariu2012}, or the object tracker employed in \textit{Slam++} \cite{SalasMoreno2013}, are compatible with the developed framework.

\subsection{Tree-like Structures}\label{ssec:m2}
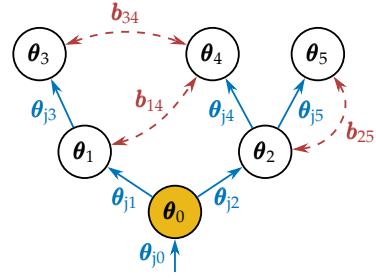
\begin{figure}[t]
	\centering

\small
\begin{tikzpicture}[line width=0.7pt]
	\node[draw, circle, fill = dlryellow] (0) at (0,0) {$\pmb{\theta}_\textrm{0}$}; 
	\node[draw, circle] (1) at (-1.2,0.8) {$\pmb{\theta}_\textrm{1}$};
	\node[draw, circle] (2) at (1.2,0.8) {$\pmb{\theta}_\textrm{2}$};
	\node[draw, circle] (3) at (-1.8,2.1) {$\pmb{\theta}_\textrm{3}$}; 
	\node[draw, circle] (4) at (0.5,2.1) {$\pmb{\theta}_\textrm{4}$};
	\node[draw, circle] (5) at (1.9,2.1) {$\pmb{\theta}_\textrm{5}$};
	\draw[-{Stealth[length=2.2mm, width=1.3mm]}, dlrblue] (0,-0.8) to node[midway, left, pos=0.5] {$\pmb{\theta}_\textrm{j0}$} (0);
	\draw[-{Stealth[length=2.2mm, width=1.3mm]}, dlrblue] (0) to node[midway, below, pos=0.6] {$\pmb{\theta}_\textrm{j1}$} (1);
	\draw[-{Stealth[length=2.2mm, width=1.3mm]}, dlrblue] (0) to node[midway, below, pos=0.6] {$~\pmb{\theta}_\textrm{j2}$} (2);
	\draw[-{Stealth[length=2.2mm, width=1.3mm]}, dlrblue] (1) to node[midway, left, pos=0.3] {$\pmb{\theta}_\textrm{j3}$} (3);
	\draw[-{Stealth[length=2.2mm, width=1.3mm]}, dlrblue] (2) to node[midway, left, pos=0.3] {$\pmb{\theta}_\textrm{j4}$} (4); 
	\draw[-{Stealth[length=2.2mm, width=1.3mm]}, dlrblue] (2) to node[midway, right, pos=0.3] {$\,\pmb{\theta}_\textrm{j5}$} (5);
	\draw[{Stealth[length=2.2mm, width=1.3mm]}-{Stealth[length=2.2mm, width=1.3mm]}, dashed, dlrred] (3) to[out=20,in=160] node[midway, above, pos=0.5] {$\pmb{b}_\textrm{34}$} (4);
	\draw[{Stealth[length=2.2mm, width=1.3mm]}-{Stealth[length=2.2mm, width=1.3mm]}, dashed, dlrred] (1) to[out=15,in=-125] node[midway, above, pos=0.5] {$\pmb{b}_\textrm{14}~~~$} (4);
	\draw[{Stealth[length=2.2mm, width=1.3mm]}-{Stealth[length=2.2mm, width=1.3mm]}, dashed, dlrred] (2) to[out=5,in=-55, distance=0.55cm] node[midway, right, pos=0.4] {~$\pmb{b}_\textrm{25}$} (5); 	
\end{tikzpicture} 
	\caption{
		Relation between variation parameters in a kinematic structure.
	 	The \ac{6DoF} pose variations $\pmb{\theta}_\textrm{0}$ to $\pmb{\theta}_\textrm{5}$ of bodies are illustrated as vertices in a graph.
	 	For the root body, the vertex is colored yellow.
	 	Connections to parent bodies and the associated joint variations $\pmb{\theta}_\textrm{j1}$ to $\pmb{\theta}_\textrm{j5}$ are indicated by blue arrows.
	 	For the root body, which does not have a parent, the joint variation $\pmb{\theta}_\textrm{j0}$ directly describes the pose variation $\pmb{\theta}_\textrm{0}$.
	 	Finally, dashed red lines indicate the constraints $\pmb{b}_\textrm{14}$, $\pmb{b}_\textrm{34}$, and  $\pmb{b}_\textrm{25}$ between bodies.
	}\label{fig:m20}
\end{figure}

In general, tree-like kinematic structures consist of individual bodies that are connected by joints.
All bodies, except for the root, have a single parent.
Based on those connections, the \ac{6DoF} pose variation $\pmb{\theta}$ of each body is described by joint variations $\pmb{\theta}_{\textrm{j}} \in \mathbb{R}^{n_\textrm{j}}$ along the kinematic chain from the root.
The number $n_\textrm{j} \in [0..6]$ describes how many degrees of freedom a particular joint allows for a body relative to its parent.
Finally, the full variation of a kinematic structure with $n$ bodies is described by a combination of all joint variations $\pmb{\theta}_\textrm{k}^\top = \begin{bmatrix} \pmb{\theta}_{\textrm{j}0}^\top \dots \pmb{\theta}_{\textrm{j}n}^\top \end{bmatrix}$.
Note that, for the root body, the joint variation $\pmb{\theta}_{\textrm{j}0}$ directly describes the pose variation $\pmb{\theta}_0$.
An example that shows the relation between individual parameters in a kinematic structure is given in Fig.\,\ref{fig:m20}.

Starting from the energy functions $E_i(\pmb{\theta}_i)$ of individual bodies, the combined energy of the kinematic structure can be calculated as the sum of the energy of all bodies
\begin{equation} \label{eq:m20}
	E_\textrm{k}(\pmb{\theta}_\textrm{k}) = \sum_{i = 1}^n E_i(\pmb{\theta}_i),
\end{equation}
where variation vectors $\pmb{\theta}_i$ of individual bodies are functions of the combined variation $\pmb{\theta}_\textrm{k}$.
Given the definitions of gradient vectors and Hessian matrices, we can now write
\begin{align} \label{eq:m21}
	\pmb{g}_\textrm{k}^\top &= \frac{\partial E_\textrm{k}}{\partial \pmb{\theta}_\textrm{k}}\bigg\vert_{\pmb{\theta}_\textrm{k}=\pmb{0}}
	= \sum_{i = 1}^n \frac{\partial E_i}{\partial \pmb{\theta}_i}\frac{\partial \pmb{\theta}_i}{\partial \pmb{\theta}_\textrm{k}}\bigg\vert_{\pmb{\theta}_\textrm{k}=\pmb{0}}
	= \sum_{i = 1}^n \pmb{g}_i^\top \pmb{J}_i,\\[5pt]
	&\begin{aligned} \label{eq:m22}
		\mathllap{\pmb{H}_\textrm{k}} = \frac{\partial^2 E_\textrm{k}}{\partial{\pmb{\theta}_\textrm{k}}^{\!2}}\bigg\vert_{\pmb{\theta}_\textrm{k}=\pmb{0}}
		&\approx \sum_{i = 1}^n \frac{\partial \pmb{\theta}_i}{\partial 	\pmb{\theta}_\textrm{k}}^{\!\top} \frac{\partial^2 E_i}{\partial {\pmb{\theta}_i}^{\!2}} \frac{\partial \pmb{\theta}_i}{\partial \pmb{\theta}_\textrm{k}}\bigg\vert_{\pmb{\theta}_\textrm{k}=\pmb{0}}\\
		&\approx \sum_{i = 1}^n \pmb{J}_i^\top \pmb{H}_i\,\pmb{J}_i,
	\end{aligned}
\end{align}
with $\pmb{g}_i$ and $\pmb{H}_i$ the gradient vector and the Hessian matrix of the rigid body $i$.
Note that second-order derivatives of pose variations $\pmb{\theta}_i$ with respect to $\pmb{\theta}_\textrm{k}$ were neglected.
The Jacobian matrices $\pmb{J}_i$ describe how the variation of the entire kinematic structure affects the pose of individual bodies.
A detailed derivation of the required body Jacobians is given in Section\,\ref{ssec:k1}.
One major difference of our formulation in comparison to the approach of Lowe \cite{Lowe1991} is that we do not calculate the change of each measurement with respect to the combined variation $\pmb{\theta}_\textrm{k}$.
Instead, the change of a body's measurements is first computed for the minimal \ac{6DoF} pose variation $\pmb{\theta}_i$ and then projected to the full kinematic structure.
This has the advantage that the same gradient vectors and Hessian matrices as for single object tracking can be used.
In addition, for kinematic structures with large numbers of bodies and measurements, it is more efficient.

\subsection{Closed Kinematic Structures}\label{ssec:m3}
To model closed kinematic structures, we start from the previously defined tree-like structures and include additional constraints.
A constraint describes any relation between bodies that can be expressed by a constraint equation $\pmb{b}(\pmb{\theta}_\textrm{k}) = \pmb{0}$.
A simple example is a rotational joint, where translational differences normal to the rotation axis have to be zero.
An illustration of the topology of a tree-like kinematic structure with constraints between individual bodies is shown in Fig.\,\ref{fig:m20}.
Given $n$ constraints, the full constraint equation for the kinematic structure is written as $\pmb{b}_\textrm{k}^\top = \begin{bmatrix} \pmb{b}_{0}^\top \dots \pmb{b}_n^\top \end{bmatrix}$.
To integrate those constraints into the combined energy function $E_\textrm{k}(\pmb{\theta}_\textrm{k})$, we write the following Lagrangian function
\begin{equation} \label{eq:m30}
	\mathcal{L}(\pmb{\theta}_\textrm{k}, \pmb{\lambda}) = E_\textrm{k}(\pmb{\theta}_\textrm{k}) + \pmb{b}_\textrm{k}(\pmb{\theta}_\textrm{k})^\top \pmb{\lambda},
\end{equation}
where $\pmb{\lambda}$ is the vector of Lagrange multipliers.
To find a solution for $\pmb{\theta}_\textrm{k}$ that satisfies the imposed constraints and minimizes the energy function, the following equation has to be solved
\begin{equation} \label{eq:m31}
	\nabla\mathcal{L}(\pmb{\theta}_\textrm{k}, \pmb{\lambda}) =
	\begin{bmatrix}
		\pmb{g}_\textrm{k}(\pmb{\theta}_\textrm{k}) + \pmb{B}_\textrm{k}(\pmb{\theta}_\textrm{k})^\top \pmb{\lambda}\\
		\pmb{b}_\textrm{k}(\pmb{\theta}_\textrm{k})
	\end{bmatrix} \overset{!}{=} \pmb{0},
\end{equation}
where the constraint Jacobian $\pmb{B}_\textrm{k}$ is the first-order derivative of $\pmb{b}_\textrm{k}$ with respect to $\pmb{\theta}_\textrm{k}$.
Detailed derivations of the constraint function and Jacobian are given in Section\,\ref{ssec:k2}.

Analogous to the minimization of the nonlinear energy function $E(\pmb{\theta})$ of a single object, we want to linearize (\ref{eq:m31}) around $\pmb{\theta}_\textrm{k} = \pmb{0}$ and iteratively find a solution.
For this, we use the first-order approximations $\pmb{g}_\textrm{k}(\pmb{\theta}_\textrm{k}) \approx \pmb{g}_\textrm{k} + \pmb{H}_\textrm{k}\pmb{\theta}_\textrm{k}$ and $\pmb{b}_\textrm{k}(\pmb{\theta}_\textrm{k}) \approx \pmb{b}_\textrm{k} + \pmb{B}_\textrm{k}\pmb{\theta}_\textrm{k}$ to derive the following linear equation
\begin{equation}\label{eq:m32}
	\renewcommand\arraystretch{1.15}
	\begin{bmatrix}
		\pmb{\hat{\theta}}_\textrm{k}\\
		\pmb{\hat{\lambda}}
	\end{bmatrix} = -
	\begin{bmatrix}
		\pmb{H}_\textrm{k} & \pmb{B}_\textrm{k}^\top \\
		\pmb{B}_\textrm{k} & \pmb{0}
	\end{bmatrix}^{-1}
	\begin{bmatrix}
		\pmb{g}_\textrm{k} \\
		\pmb{b}_\textrm{k}
	\end{bmatrix}.
\end{equation}
Similar to (\ref{eq:m10}), the derived linear equation is used to estimate $\pmb{\hat{\theta}}_\textrm{k}$ and to iteratively minimize $E_\textrm{k}(\pmb{\theta}_\textrm{k})$ while, at the same time, considering the constraints $\pmb{b}_\textrm{k}(\pmb{\theta}_\textrm{k}) = \pmb{0}$.
Note that, in general, given a positive definite Hessian $\pmb{H}_\textrm{k}$ and unique and non-contradicting constraint equations that lead to non-zero and linearly independent row vectors in the constraint Jacobian $\pmb{B}_\textrm{k}$, the matrix in (\ref{eq:m32}) is invertible and a unique solution exists for the estimated variation vector $\pmb{\hat{\theta}}_\textrm{k}$.

Finally, in contrast to the \textit{postimposed} method of Drummond and Cipolla \cite{Drummond2002b}, which first predicts unconstrained poses and applies constraints later, our formulation directly estimates a variation vector that is compatible with constraints.
In addition, the approach is not limited to velocity constraints.
Together with our formulation for tree-like kinematic structures, the developed multi-body tracking framework supports an efficient minimal parameterization while simultaneously providing the flexibility to add constraints for closed chains and temporary limitations.
It, therefore, combines the best of both worlds.


\section{Parameterization}\label{sec:k}
In this section, we derive all equations that are required to define a kinematic structure.
For the pose parameterization, we thereby focus on the axis-angle representation.
Note that while tracking algorithms based on Euler angles, twist coordinates, or quaternions have also been developed, in our experience, this parameterization is the most popular.
After an introduction to general mathematical concepts, Jacobian matrices and constraint equations are derived.
In both cases, we allow to lock or release the rotation and translation along individual coordinate axes, facilitating various different joints and constraints.
Finally, we describe how to update the pose of individual bodies.

\subsection{Preliminaries}\label{ssec:k0}
To define the pose between two reference frames $\textrm{A}$ and $\textrm{B}$, the following transformation matrix ${}_\textrm{A}\pmb{T}_\textrm{B}\in \mathbb{SE}(3)$ is used
\begin{equation} \label{eq:k00}
	{}_\textrm{A}\pmb{T}_\textrm{B} =
	\begin{bmatrix}
		_\textrm{A}\pmb{R}_\textrm{B} & _\textrm{A}\pmb{t}_\textrm{B} \\ \pmb{0} & 1
	\end{bmatrix},
\end{equation}
where $_\textrm{A}\pmb{R}_\textrm{B} \in \mathbb{SO}(3)$ is a rotation matrix and $_\textrm{A}\pmb{t}_\textrm{B} \in \mathbb{R}^3$ is a translation vector.
They together describe the transformation from $\textrm{B}$ to $\textrm{A}$.
For the variation of poses with a minimal set of parameters, we adopt the axis-angle representation for the rotation and write
\begin{equation} \label{eq:k01}
	\pmb{T}(\pmb{\theta}) =
	\begin{bmatrix}
		\exp([\pmb{\theta}_\textrm{r}]_\times) & \pmb{\theta}_\textrm{t} \\ \pmb{0} & 1
	\end{bmatrix},
\end{equation}
where $\pmb{\theta}_\textrm{r}\in \mathbb{R}^3$ and $\pmb{\theta}_\textrm{t}\in \mathbb{R}^3$ are the rotational and translational components of the variation vector $\pmb{\theta}^\top = \begin{bmatrix} \pmb{\theta}_\textrm{r}^\top & \pmb{\theta}_\textrm{t}^\top \end{bmatrix}$, and $[\pmb{\theta}_\textrm{r}]_\times$ is the skew-symmetric cross-product matrix of $\pmb{\theta}_\textrm{r}$.
Finally, to project the variation vector $\pmb{\theta}$ from reference frame $\textrm{B}$ to $\textrm{A}$, the following adjoint representation is used
\begin{equation} \label{eq:k02}
	\Ad({}_\textrm{A}\pmb{T}_\textrm{B}) =
	\begin{bmatrix}
		{}_\textrm{A}\pmb{R}_\textrm{B} & \pmb{0} \\
		[{}_\textrm{A}\pmb{t}_\textrm{B}]_\times\,{}_\textrm{A}\pmb{R}_\textrm{B} & {}_\textrm{A}\pmb{R}_\textrm{B}
	\end{bmatrix}.
\end{equation}
Based on those general mathematical concepts, we are able to describe kinematic structures and derive the required body Jacobians and constraint equations.

\subsection{Body Jacobians}\label{ssec:k1}
\begin{figure}[t]
	\centering
	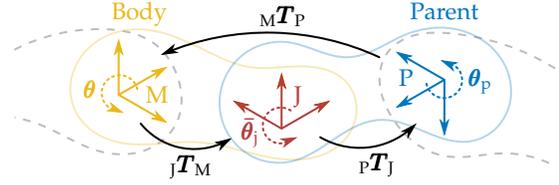
	\caption{
		Visualization of a body that is connected to its parent by a joint.
		Model, parent, and joint coordinate frames $\textrm{M}$, $\textrm{P}$, and $\textrm{J}$ are illustrated in yellow, blue, and red, respectively.
		For the coordinate frames, the corresponding variation vectors $\pmb{\theta}$, $\pmb{\bar{\theta}}_\textrm{j}$, and $\pmb{\theta}_\textrm{p}$, as well as the relative homogeneous transformations ${}_\textrm{J}\pmb{T}_\textrm{M}$, ${}_\textrm{P}\pmb{T}_\textrm{J}$, and ${}_\textrm{M}\pmb{T}_\textrm{P}$, are shown.
	}\label{fig:k10}
\end{figure}

Each body has a defined location relative to its joint and corresponding parent.
In the following, we introduce a general class of joints that allow motion along a user-defined set of rotational and translational axes.
Based on the joint reference frame $\textrm{J}$, motion along these directions is modeled by the variation vector $\pmb{\theta}_\textrm{j}$.
The full \ac{6DoF} pose variation $\pmb{\theta}$ of a body's model frame $\textrm{M}$ given the variation of its parent and joint is then defined as follows
\begin{equation}\label{eq:k10}
	\pmb{T}(\pmb{\theta}) = {}_\textrm{M}\pmb{T}_\textrm{P}\, \pmb{T}(\pmb{\theta}_\textrm{p})\,
	{}_\textrm{P}\pmb{T}_\textrm{J}\, \pmb{T}(\pmb{\bar{\theta}}_\textrm{j})\,  
	{}_\textrm{J}\pmb{T}_\textrm{M}, 
\end{equation}
where $\textrm{P}$ is the reference frame of the parent and $\pmb{\theta}_\textrm{p}$ the corresponding \ac{6DoF} variation vector.
An illustration of the transformations is shown in Fig.\,\ref{fig:k10}.
The bar above the joint variation vector $\pmb{\bar{\theta}}_\textrm{j}$ indicates the extension of $\pmb{\theta}_\textrm{j}$, which has $n_\textrm{j}$ dimensions, to a \ac{6DoF} vector.
For a typical joint that only allows movement along specified axes of the joint frame, we simply set fixed directions to zero.

To project equations of individual bodies to the multi-body system, Jacobians are used.
Given a body's joint and parent, we are able to formulate a recursive strategy to calculate the Jacobian.
Based on the Jacobian of the parent body $\pmb{J}_\textrm{p}$, we calculate
\begin{equation}\label{eq:k11}
	\pmb{J} = \frac{\partial \pmb{\theta}}{\partial \pmb{\theta}_\textrm{k}}\bigg\vert_{\pmb{\theta}_\textrm{k}=\pmb{0}}
	= \frac{\partial \pmb{\theta}}{\partial \pmb{\theta}_\textrm{p}} \pmb{J}_\textrm{p} + 
	\begin{bmatrix}
		\pmb{0} &\frac{\partial \pmb{\theta}}{\partial \pmb{\theta}_\textrm{j}} &\pmb{0}
	\end{bmatrix}\bigg\vert_{\pmb{\theta}_\textrm{k}=\pmb{0}},
\end{equation}
where the partial derivative $\frac{\partial \pmb{\theta}}{\partial \pmb{\theta}_\textrm{j}}$ has to be added at the correct location that corresponds to the position of $\pmb{\theta}_\textrm{j}$ in the vector $\pmb{\theta}_\textrm{k}$.
The advantage of this recursive strategy is that it is not necessary to assemble the derivatives of all individual joints along the kinematic chain.
Instead, we simply consider the kinematic chain using the Jacobian of the parent.
For the root body, the parent Jacobian $\pmb{J}_\textrm{p}$ is simply zero.
To calculate the Jacobian, we now only require the partial derivatives of the body variation with respect to the variations of the joint and parent.

Starting from (\ref{eq:k10}) and knowing that the adjoint representation in (\ref{eq:k02}) can be used to project variation vectors between reference frames, we write the following relations
\begin{align}\label{eq:k12}
	\pmb{\theta} \vert_{\pmb{\theta}_\textrm{j}=\pmb{0}}
	&=\Ad({}_\textrm{M}\pmb{T}_\textrm{P})\,\pmb{\theta}_\textrm{p},\\[5pt]\label{eq:k13}
	\pmb{\theta} \vert_{\pmb{\theta}_\textrm{p}=\pmb{0}}
	&=\Ad({}_\textrm{M}\pmb{T}_\textrm{J})\,\pmb{\bar{\theta}}_\textrm{j}.
\end{align}
As can be seen from those equations, the required first-order derivatives are
\begin{align}\label{eq:k14}
	\frac{\partial \pmb{\theta}}{\partial \pmb{\theta}_\textrm{p}} \bigg\vert_{\pmb{\theta}_\textrm{k}=\pmb{0}}
	&=\Ad({}_\textrm{M}\pmb{T}_\textrm{P}),\\[5pt]\label{eq:k15}
	\frac{\partial \pmb{\theta}}{\partial \pmb{\bar{\theta}}_\textrm{j}} \bigg\vert_{\pmb{\theta}_\textrm{k}=\pmb{0}}
	&=\Ad({}_\textrm{M}\pmb{T}_\textrm{J}).
\end{align}
Note that (\ref{eq:k15}) considers the derivative with respect to the extended variation vector $\pmb{\bar{\theta}}_\textrm{j}$.
To get the first-order derivative $\frac{\partial \pmb{\theta}}{\partial \pmb{\theta}_\textrm{j}}\big\vert_{\pmb{\theta}_\textrm{k}=\pmb{0}}$ with respect to the variation vector $\pmb{\theta}_\textrm{j}$, one simply assembles the columns from the partial derivative in (\ref{eq:k15}) that are considered by $\pmb{\theta}_\textrm{j}$.
With the derived formulation, we are able to model various joints, including revolute, prismatic, or spherical joints.
For more specific joints, such as screw connections, it is typically also straightforward to derive the necessary first-order derivatives as a combination of the motion along individual axes.
Instead of directly assembling columns from (\ref{eq:k15}), one then simply multiplies with the joint's partial derivative $\frac{\partial\pmb{\bar{\theta}}_\textrm{j}}{\partial\pmb{\theta}_\textrm{j}}$ to get the derivative $\frac{\partial\pmb{\theta}}{\partial\pmb{\theta}_\textrm{j}} = \frac{\partial\pmb{\theta}}{\partial\pmb{\bar{\theta}}_\textrm{j}}\frac{\partial\pmb{\bar{\theta}}_\textrm{j}}{\partial\pmb{\theta}_\textrm{j}}$ required for the body Jacobian.

\subsection{Constraint Equations}\label{ssec:k2}
\begin{figure}[t]
	\centering
	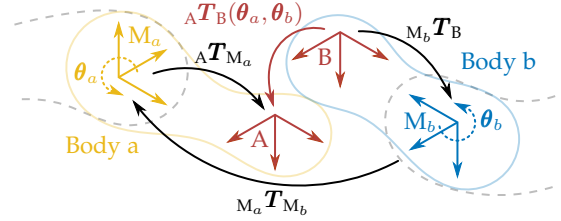
	\caption{
		Illustration of a constraint between two bodies $a$ and $b$.
		The model coordinate frames $\textrm{M}_a$ and $\textrm{M}_b$, as well as the variation vectors $\pmb{\theta}_a$ and $\pmb{\theta}_b$ are illustrated in yellow and blue, respectively.
		The constraint coordinate frames $A$ and $B$ associated with bodies $a$ and $b$ are colored in red.
		The relative transformation ${}_\textrm{A}\pmb{T}_\textrm{B}(\pmb{\theta}_a, \pmb{\theta}_b)$ is calculated from ${}_{\textrm{M}_b}\pmb{T}_\textrm{B}$, ${}_{\textrm{M}_a}\pmb{T}_{\textrm{M}_b}$, and ${}_\textrm{A}\pmb{T}_{\textrm{M}_a}$ and is considered in the constraint equation. 
	}\label{fig:k20}
\end{figure}

In the following, we derive constraints that consider rotational and translational differences between two bodies along individual axes.
Differences are calculated with respect to the constraint reference frames $\textrm{A}$ and $\textrm{B}$, which are defined for the corresponding bodies $a$ and $b$.
Constrained differences between those two frames are enforced to be zero.
The relative pose between frames $\textrm{A}$ and $\textrm{B}$, given the pose variation vectors $\pmb{\theta}_a$ and $\pmb{\theta}_b$, is written as
\begin{equation}\label{eq:k20}
	{}_{\textrm{A}}\pmb{T}_{\textrm{B}}(\pmb{\theta}_a, \pmb{\theta}_b) = 
	{}_{\textrm{A}}\pmb{T}_{\textrm{M}_a}\, \pmb{T}(\pmb{\theta}_a)^{-1}\,
	{}_{\textrm{M}_a}\pmb{T}_{\textrm{M}_b}\, \pmb{T}(\pmb{\theta}_b)\,
	{}_{\textrm{M}_b}\pmb{T}_{\textrm{B}},
\end{equation}
where $\textrm{M}_a$ and $\textrm{M}_b$ are the model frames of bodies $a$ and $b$.
An illustration of the transformation is shown in Fig.\,\ref{fig:k20}.
Based on the relative pose ${}_{\textrm{A}}\pmb{T}_{\textrm{B}}(\pmb{\theta}_a, \pmb{\theta}_b)$, we define the extended \ac{6DoF} constraint equation for all rotational and translational axes as follows
\begin{equation}\label{eq:k21}
	\pmb{\bar{b}}_{ab}(\pmb{\theta}_a, \pmb{\theta}_b) = 
	\begin{bmatrix}
		{}_{\textrm{A}}\pmb{r}_{\textrm{B}}(\pmb{\theta}_a, \pmb{\theta}_b)\\
		{}_{\textrm{A}}\pmb{t}_{\textrm{B}}(\pmb{\theta}_a, \pmb{\theta}_b)
	\end{bmatrix},
\end{equation}
where ${}_{\textrm{A}}\pmb{r}_{\textrm{B}}$ is the rotation vector, and ${}_{\textrm{A}}\pmb{t}_{\textrm{B}}$ is the translation vector of the transformation matrix ${}_{\textrm{A}}\pmb{T}_{\textrm{B}}$.
The rotation vector and the rotation matrix are related by the exponential map ${}_{\textrm{A}}\pmb{R}_{\textrm{B}} = \exp([{}_{\textrm{A}}\pmb{r}_\textrm{B}]_\times)$.
Note that, depending on the directions that should be constrained, only individual elements of the extended constraint equation $\pmb{\bar{b}}_{ab}$ are used.
The resulting reduced constraint equation is denoted $\pmb{b}_{ab}$.

Given a constraint equation, the corresponding constraint Jacobian can be calculated using body Jacobians $\pmb{J}$ that were derived in the previous section as follows
\begin{equation}\label{eq:k22}
	\pmb{B}_{ab} =
	\frac{\partial \pmb{b}_{ab}}{\partial \pmb{\theta}_\textrm{k}} \bigg\vert_{\pmb{\theta}_\textrm{k}=\pmb{0}}
	= \frac{\partial\pmb{b}_{ab}}{\partial\pmb{\theta}_a}\pmb{J}_a + \frac{\partial \pmb{b}_{ab}}{\partial \pmb{\theta}_b}\pmb{J}_b
	\bigg\vert_{\pmb{\theta}_\textrm{k}=\pmb{0}}.
\end{equation}
For the required partial derivatives of the extended constraint equation $\pmb{\bar{b}}_{ab}$ with respect to the pose variations $\pmb{\theta}_a$ and $\pmb{\theta}_b$, the calculation is quite lengthy.
This is especially the case for the rotational constraint, which considers the nontrivial change of the rotation vector ${}_\textrm{A}\pmb{r}_\textrm{B}$ for the variation of the two bodies.
Consequently, we provide the full derivation in Appendix\,\ref{sec:a0} and only state the following final results
\begin{align}\label{eq:k23}
	\frac{\partial\pmb{\bar{b}}_{ab}}{\partial\pmb{\theta}_a}\bigg\vert_{\pmb{\theta}_\textrm{k}=\pmb{0}}
	&=
	\begin{bmatrix}
		-\pmb{C}\,{}_\textrm{A}\pmb{R}_{\textrm{M}_a} & \pmb{0}\\
		{}_\textrm{A}\pmb{R}_{\textrm{M}_a}\,[{}_{\textrm{M}_a}\pmb{t}_\textrm{B}]_\times & -{}_\textrm{A}\pmb{R}_{\textrm{M}_a}
	\end{bmatrix},\\[5pt]\label{eq:k24}
	\frac{\partial\pmb{\bar{b}}_{ab}}{\partial\pmb{\theta}_b}\bigg\vert_{\pmb{\theta}_\textrm{k}=\pmb{0}}
	&=
	\begin{bmatrix}
		\pmb{C}\, {}_\textrm{A}\pmb{R}_{\textrm{M}_b} & \pmb{0}\\
		-{}_\textrm{A}\pmb{R}_{\textrm{M}_b}\,[{}_{\textrm{M}_b}\pmb{t}_\textrm{B}]_\times & {}_\textrm{A}\pmb{R}_{\textrm{M}_b}
	\end{bmatrix}.
\end{align}
To obtain the derivatives for the reduced constraint equation $\pmb{b}_{ab}$, which are required in (\ref{eq:k22}), one simply assembles the rows from (\ref{eq:k23}) and (\ref{eq:k24}) that correspond to the constrained axes of coordinate frame $\textrm{A}$.
Finally, given the Jacobians $\pmb{B}_{ab}$ of individual constraints, the full constraint Jacobian $\pmb{B}_\textrm{k}$ can be assembled by row-wise concatenation of those matrices.

The matrix $\pmb{C}$, which is used in the calculated first-order derivatives and which we call variation matrix, is defined as
\begin{equation}\label{eq:k25}
	\pmb{C} = \frac{\alpha}{2} \cot\Big(\frac{\alpha}{2}\Big)\pmb{I}-\frac{\alpha}{2}[\pmb{e}]_\times+\Big(1-\frac{\alpha}{2}\cot\Big(\frac{\alpha}{2}\Big)\Big)\pmb{e}\pmb{e}^\top,
\end{equation}
where $\alpha\in\mathbb{R}$ is the rotation angle and $\pmb{e}\in\mathbb{R}^3$ with $\lVert\pmb{e}\rVert_2 = 1$ is the rotation axis of the rotation vector ${}_\textrm{A}\pmb{r}_\textrm{B} = \alpha \pmb{e}$ evaluated at $\pmb{\theta}_\textrm{k}=\pmb{0}$.
The matrix describes how the rotation vector ${}_\textrm{A}\pmb{r}_\textrm{B}$ changes with the variation of a subsequent infinitesimal rotation.
It is derived in Appendix\,\ref{sec:a0} as part of the calculation of (\ref{eq:k23}) and (\ref{eq:k24}).
In Appendix\,\ref{sec:a1}, we show that the variation matrix $\pmb{C}$ is a normal, non-symmetric, and non-orthogonal matrix that is directly related to the rotation matrix by
\begin{equation}
	{}_\textrm{A}\pmb{R}_\textrm{B} = \pmb{C}^\top \pmb{C}^{-1} = \pmb{C}^{-1}\pmb{C}^\top.
\end{equation}
Also, for the rotation vector ${}_\textrm{A}\pmb{r}_\textrm{B}$, we are able to prove that
\begin{equation}
	{}_\textrm{A}\pmb{r}_\textrm{B} = \pmb{C}\,{}_\textrm{A}\pmb{r}_\textrm{B} =  \pmb{C}^\top{}_\textrm{A}\pmb{r}_\textrm{B}.
\end{equation}
This shows that the rotation vector ${}_\textrm{A}\pmb{r}_\textrm{B}$ is an eigenvector of both the original and transposed variation matrix $\pmb{C}$.

For the approach in Section\,\ref{ssec:m2}, which uses body Jacobians to project equations to a minimal parameterization, only exact solutions are possible.
At the same time, for the constraint approach in Section\,\ref{ssec:m3} and the equations proposed in this section, kinematic compliance is not apparent.
Especially for the rotation, which is highly non-linear in Euclidean space, one could expect that multiple iterations of the Newton optimization are required to converge to a kinematically accurate result.
However, in Appendix\,\ref{sec:a3}, we are able to prove mathematically that this is not the case.
In our proof, the developed constraint equations and Jacobians are introduced into the linear equations of the Newton optimization.
Also, the previously developed property that $	{}_\textrm{A}\pmb{r}_\textrm{B} = \pmb{C}{}_\textrm{A}\pmb{r}_\textrm{B}$ is used.
Based on this, we calculate the relative pose change per iteration.
The results demonstrate that both rotational and translational constraints converge to a kinematically compliant result in a single iteration.
While the proof in Appendix\,\ref{sec:a3} assumes that constraint coordinate frames are equal to model frames, experiments in Section\,\ref{ssec:e2} demonstrate that results also hold for more general cases.

If constraint equations are fulfilled and $\pmb{\bar{b}}_{ab} = \pmb{0}$, the coordinate frames $\textrm{A}$ and $\textrm{B}$ are equal, and the variation matrix $\pmb{C}$ reduces to the identity matrix.
As a consequence, the derivatives in (\ref{eq:k23}) and (\ref{eq:k24}) turn into the adjoint representations $-\Ad({}_\textrm{A}\pmb{T}_{\textrm{M}_a})$ and $\Ad({}_\textrm{A}\pmb{T}_{\textrm{M}_b})$.
A short proof of this is given in Appendix\,\ref{sec:a2}.
The derived expressions directly correspond to the formulation of Drummond and Cipolla \cite{Drummond2002b}.
This equivalence demonstrates that our approach is an extension of their method.
However, the big advantage of our generalization is that we directly operate on pose differences instead of velocities.
This ensures that even if the pose constraint is not fulfilled initially, our method will automatically converge to a consistent result.
As a consequence, it is possible to model closed chains or add temporary constraints that are not currently fulfilled.
Both cases are not viable using velocity constraints.
In addition, we would like to highlight the benefits of directly using the rotation vector in the constraint equations.
Compared to other options, such as using orthogonal axes, our formulation has the advantage that it provides consistent results over the entire space of possible rotations.
We are, therefore, able to successfully recover even from maximum rotational differences of $\alpha=\pi$ in a single iteration.

\subsection{Pose Update}\label{ssec:k3}
Given the derived Jacobians and constraint equations, we are able to estimate the variation vector of the kinematic structure $\pmb{\theta}_\textrm{k}^\top = \begin{bmatrix} \pmb{\theta}_{\textrm{j}0}^\top \dots \pmb{\theta}_{\textrm{j}n}^\top \end{bmatrix}$ using the linear relation of the Newton optimization in (\ref{eq:m32}).
Similar to the calculation of body Jacobians, we formulate a recursive strategy for the pose update.
With the estimated joint variation $\pmb{\hat{\theta}}_\textrm{j}$ and the pose of the parent, the body pose is updated as follows
\begin{equation}
	{}_\textrm{A}\pmb{T}_\textrm{M}^+ = {}_\textrm{A}\pmb{T}_\textrm{P}\, {}_\textrm{P}\pmb{T}_\textrm{J}\, \pmb{T}(\pmb{\hat{\bar{\theta}}}_\textrm{j})\, {}_\textrm{J}\pmb{T}_\textrm{M}, 
\end{equation}
where $\textrm{A}$ is some arbitrary frame of reference such as the camera frame.
Like in (\ref{eq:k10}), the full \ac{6DoF} variation vector $\pmb{\hat{\bar{\theta}}}_\textrm{j}$ has to be used for the calculation of the transformation matrix.
Again, it is simply created by adding zeros to the fixed axes of our joint variation.
Note that for the pose of the joint coordinate frame $\textrm{J}$, either the transformation with respect to the body ${}_\textrm{J}\pmb{T}_\textrm{M}$ or the parent ${}_\textrm{P}\pmb{T}_\textrm{J}$ is fixed.
The other is inferred from the previous pose estimates of the body and parent, as well as the fixed transformation.
While most of the time, both options are valid, for some joints, which allow, for example, rotation around two axes, it makes a difference if ${}_\textrm{J}\pmb{T}_\textrm{M}$ or ${}_\textrm{P}\pmb{T}_\textrm{J}$ is fixed.

Finally, to update the root body, which does not have a parent, the following strategy is adopted
\begin{equation}
	{}_\textrm{A}\pmb{T}_{\textrm{M}_0}^+ = {}_\textrm{A}\pmb{T}_{\textrm{M}_0}\,
	{}_{\textrm{M}_0}\pmb{T}_{\textrm{J}_0}\, \pmb{T}(\pmb{\hat{\bar{\theta}}}_{\textrm{j}0})\,	{}_{\textrm{J}_0}\pmb{T}_{\textrm{M}_0}.
\end{equation}
It considers the variation with respect to the previous pose estimate in the joint's frame of reference.
In summary, the developed approach allows us to estimate all body poses in the kinematic structure, ensuring that variation vectors comply with constraints and poses are only updated along free joint directions.


\section{Implementation}\label{sec:i}
The following section considers the extension of our work \textit{ICG} \cite{Stoiber2022} to multi-body tracking.
\textit{ICG} is a highly-efficient probabilistic tracker that considers sparse region and depth information for the tracking of multiple independent objects.
For an efficient representation of geometry that avoids online rendering, it utilizes a so-called sparse viewpoint model that stores randomly sampled contour and surface points, as well as other information, for views all around the object.
For the region modality, color histograms are used to differentiate between the object and the background.
During tracking, contour points are projected into the image to define the location of so-called correspondence lines.
For each line, a discrete probability distribution for the contour location is computed based on the probability that pixels belong to the foreground or background.
For the depth modality, an \textit{ICP}-like approach is implemented that computes the distance between surface model points and closest correspondence points from the depth image.
It uses the point-to-plane error metric and assumes a normal distribution.
For the maximization of the joint \ac{PDF}, \textit{ICG} employs Newton optimization with Tikhonov regularization.
To parameterize the pose variation, it adopts the axis-angle representation.
\textit{ICG}, therefore, fits perfectly into the developed multi-body tracking framework.
For the implementation, we directly build on the available source code\footnote{\url{https://github.com/DLR-RM/3DObjectTracking}}.
If not stated otherwise, parameter values from the corresponding publication \cite{Stoiber2022} are used.
In the following, we describe modifications that are required for the extension to multi-body tracking.

\begin{figure}[t]
	\vspace{1mm}
	\centering
\begingroup%
  \makeatletter%
  \providecommand\color[2][]{%
    \errmessage{(Inkscape) Color is used for the text in Inkscape, but the package 'color.sty' is not loaded}%
    \renewcommand\color[2][]{}%
  }%
  \providecommand\transparent[1]{%
    \errmessage{(Inkscape) Transparency is used (non-zero) for the text in Inkscape, but the package 'transparent.sty' is not loaded}%
    \renewcommand\transparent[1]{}%
  }%
  \providecommand\rotatebox[2]{#2}%
  \newcommand*\fsize{\dimexpr\f@size pt\relax}%
  \newcommand*\lineheight[1]{\fontsize{\fsize}{#1\fsize}\selectfont}%
  \ifx\svgwidth\undefined%
    \setlength{\unitlength}{238.11023622bp}%
    \ifx\svgscale\undefined%
      \relax%
    \else%
      \setlength{\unitlength}{\unitlength * \real{\svgscale}}%
    \fi%
  \else%
    \setlength{\unitlength}{\svgwidth}%
  \fi%
  \global\let\svgwidth\undefined%
  \global\let\svgscale\undefined%
  \makeatother%
  \begin{picture}(1,0.44047619)%
    \lineheight{1}%
    \setlength\tabcolsep{0pt}%
    \put(0.62561088,0.27664862){\makebox(0,0)[lt]{\lineheight{1.25}\smash{\begin{tabular}[t]{l}\small\end{tabular}}}}%
    \put(0,0){\includegraphics[width=\unitlength,page=1]{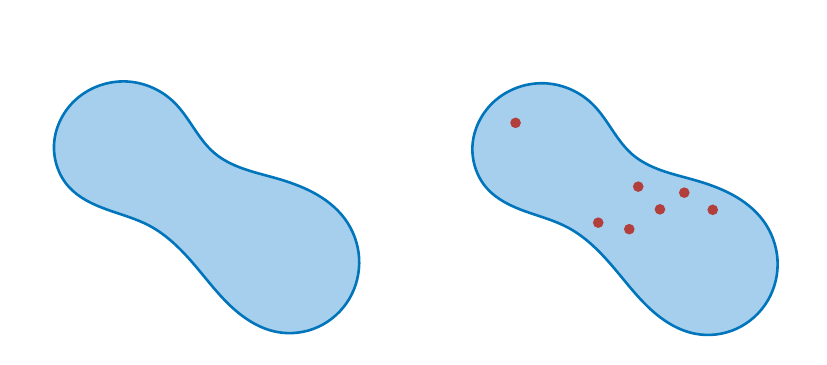}}%
    \put(0.25,0.39285714){\color[rgb]{0,0,0}\makebox(0,0)[t]{\lineheight{1.25}\smash{\begin{tabular}[t]{c}\small Region Modality\end{tabular}}}}%
    \put(0.75,0.39285714){\color[rgb]{0,0,0}\makebox(0,0)[t]{\lineheight{1.25}\smash{\begin{tabular}[t]{c}\small Depth Modality\end{tabular}}}}%
    \put(0,0){\includegraphics[width=\unitlength,page=2]{validation.pdf}}%
  \end{picture}%
\endgroup%

	\caption{
		Validation strategies for the region and depth modality.
		Valid points are visualized with a circle, while invalid points are marked by a cross.
		For the region modality, the body in the center and lower right corner are rendered with the same region ID.
		To check if a contour point is valid, pixels on a line along the normal vector are evaluated.
		If the distance to the inside is interrupted by another silhouette or the distance to the outside is interrupted by the same silhouette, the corresponding contour point is rejected.	
		For the depth modality, all bodies are rendered with unique values.
		Any surface point that is not on the body's own silhouette is invalid.
	}\label{fig:i00}
\end{figure}

For kinematic structures, it is very common that individual bodies intersect or that they are very close together.
Also, different bodies often belong to the same region.
As a consequence, one of the main modifications of \textit{ICG} considers the validation of contour and surface points.
To validate surface points from depth modalities, a silhouette image is rendered that contains all bodies.
Based on this image, surface points of a body are considered valid if they fall on the corresponding silhouette.
Note that the same rendering can be employed for multiple modalities that use the same camera.
For contour points from region modalities, we also render a silhouette image.
However, instead of using a unique ID for each body, a region ID is assigned to each body, which is rendered to the respective silhouette.
Bodies with the same region ID belong to the same region and consider similar color statistics.
As described in \cite{Stoiber2022}, it is essential that, for each correspondence line, the body region on the inside and the non-body region on the outside of the contour are not interrupted for a minimum continuous distance along the line, which was defined as $3$ segments.
Consequently, for each contour point, we check if rendered pixels along the normal vector fulfill this rule.
If it is violated, the contour point is rejected.
An illustration of the described validation strategies is shown in Fig.~\ref{fig:i00}.
In addition to contour point validation, the silhouette image is also used to check whether pixel values that should be assigned to color histograms are on the correct silhouette.
If they are on the wrong silhouette, they are ignored.
Both the rendering and validation are repeated for each correspondence search.

In addition to validation strategies, we allow individual region modalities to share color histograms.
This is especially beneficial if individual bodies are occluded while other bodies that model the same region are still visible.
Also, color histograms consider more information and, therefore, perform slightly better.
Another modification required for multi-body tracking considers the number of correspondence lines and points each modality deploys.
Since the size of bodies varies widely, the number has to be dynamically adapted to uniformly cover the contour and surface of all bodies.
We, therefore, scale the number of lines and points according to the surface area and contour length of a body's current silhouette relative to a reference area and length.
In the evaluation, we use the maximum contour length and surface area of all views and bodies within an object as reference.
Also, except for the \textit{Medical Robot}, which is shown in Fig.~\ref{fig:e01}, a maximum of $300$ correspondence lines and points is deployed per body.
For the \textit{Medical Robot}, which consists of equally sized bodies, we only use $100$ lines and points to improve computational efficiency.

The last modification considers the maximization of the joint probability using the Newton method.
Similar to \textit{ICG}, the required variation vector  $\hat{\pmb{\theta}}_\textrm{k}$ is estimated using the linear relation in (\ref{eq:m32}).
However, in contrast to the original implementation, we do not add regularization to each body separately.
Instead, we directly add the diagonal regularization matrix $\pmb{\Sigma}$ to the Hessian of the kinematic structure $\pmb{H}_\textrm{k}$.
The diagonal elements of $\pmb{\Sigma}$ consist of the rotational and translational parameters $\lambda_\textrm{r} = 100$ and $\lambda_\textrm{t} = 1000$.
They are applied to the respective rotational and translational components of the full variation vector $\pmb{\theta}_\textrm{k}$.
Finally, the constraint Jacobian $\pmb{B}_\textrm{k}$ and the Hessian matrix $\pmb{H}_\textrm{k}$ have very different orders of magnitude.
To ensure stability when solving the linear equation of the optimization, we use a robust Cholesky decomposition with pivoting.

Since, in the evaluation, our tracker deals with large frame-to-frame pose differences, we conduct $6$ iterations in which correspondences are established.
For the scale parameter of the region modality, we use $s=\{9, 7, 5, 2\}$.
Also, the correspondence point threshold of the depth modality, which is given in millimeters, is defined as $r_\textrm{t} = \{100, 80, 50\}$, and is sampled with a stride of $8\,\unit{mm}$.
Note that values for the thresholds $r_\textrm{t}$ and the stride are scaled according to the depth of model points and are defined at a reference distance of $1\,\unit{m}$.
In addition to those parameters, we use the standard deviations $\sigma_\textrm{r} = \{25, 15, 10\}$ and $\sigma_\textrm{d} = \{50, 30, 20\}$, which are given in pixels and millimeters, respectively.
The provided sets define parameters for each iteration, with the last value employed in all remaining iterations.
Finally, with the developed framework and the described modifications, our \textit{Multi-body ICG} algorithm can be used for the tracking of a wide variety of open and closed kinematic structures.


\section{Evaluation}\label{sec:e}
In the following section, we provide a detailed evaluation of the developed framework and \textit{Multi-body ICG} tracker.
For this, a newly created dataset and the used evaluation metrics are introduced.
Based on those sequences, we assess different configurations of kinematic structures with respect to quality and efficiency.
After this, the convergence of constraints is analyzed.
In addition, while the focus of this paper is a general framework that can be used with various methods, a comparison to another algorithm is also provided.
Finally, we discuss limitations of our approach.
All experiments are conducted on a computer with an \textit{Intel Core i9-11900K} CPU and a \textit{NVIDIA RTX A5000} GPU.
Qualitative results in real-world applications, as well as sequences from our dataset, are provided in the supplementary video.\footnote{\url{https://www.youtube.com/watch?v=0ORZvDDbDjA}}
Note that both the source code for our tracker and the dataset are publicly available.\textsuperscript{\ref{foo:dataset},\ref{foo:code}}

\subsection{Robot Tracking Benchmark}\label{ssec:e0}
In the past, evaluations of multi-body tracking and pose estimation methods considered only a very limited number of sequences.
Consequently, experiments were typically of a more qualitative nature \cite{Krainin2011, Klingensmith2013, Bohg2014, Michel2015, Pauwels2015, Schmidt2015a}.
The main reason is that diverse real-world data with moving cameras and high-quality pose annotations is hard to obtain, even if object configurations can be measured.
However, for a reliable evaluation, a sufficient number of realistic sequences with accurate ground truth is essential.

\begin{figure}[t]
	\centering
	\input{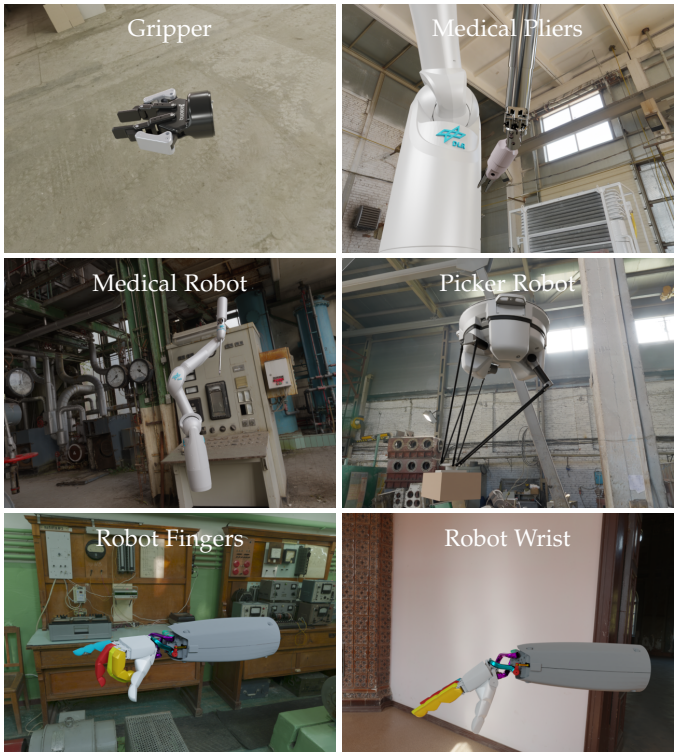}
	\caption{
		Multi-body objects included in the \textit{\ac{RTB}} dataset, featuring the \textit{Gripper}, \textit{Medical Pliers}, \textit{Medical Robot}, \textit{Picker Robot}, \textit{Robot Fingers}, and \textit{Robot Wrist}.
		The \textit{Gripper}, \textit{Picker Robot}, and \textit{Robot Wrist} contain closed kinematic chains.
		Note that for the \textit{Robot Fingers}, the wrist is locked, and only finger joints are moving, while the \textit{Robot Wrist} features locked fingers and a movable wrist.
	}\label{fig:e01}
\end{figure}

\begin{figure}[t]
	\centering
	\input{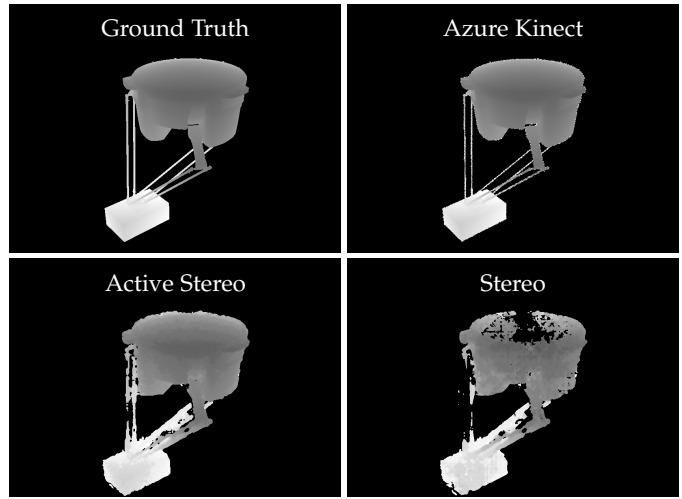}
	\caption{
		Depth image qualities provided in the \textit{\ac{RTB}} dataset, featuring \textit{Ground Truth} with perfect measurements, noise characteristics from an \textit{Azure Kinect} camera, and stereo reconstructions from \textit{SGM}.
		For \textit{Active Stereo}, the pattern from a dot projector is simulated to add additional texture information, while \textit{Stereo} operates without a pattern.
	}\label{fig:e02}
\end{figure}

We therefore created the \textit{\acf{RTB}}.
It is a novel synthetic dataset that was developed using the procedural rendering pipeline \textit{BlenderProc}\footnote{\url{https://github.com/DLR-RM/BlenderProc}}.
Example images of the dataset are shown in Fig.\,\ref{fig:e01}.
To facilitate the creation of the dataset, we extended \textit{BlenderProc} to load robot models from \textit{URDF} files.
Also, for the generation of articulated motions, forward and backward kinematics were integrated.
Our open-source pipeline produces photo-realistic sequences with \textit{HDRi} lighting\footnote{\url{https://polyhaven.com/hdris}} and physically-based materials.
Perfect ground truth annotations for camera and robot trajectories are provided in the \textit{BOP} format \cite{Hodan2020}.
Many physical effects, such as motion blur, rolling shutter, and camera shaking, are accurately modeled to reflect real-world conditions.
For each frame, we also generated the following four depth qualities: \textit{Ground Truth}, \textit{Azure Kinect}, \textit{Active Stereo}, and \textit{Stereo}.
They simulate sensors with different characteristics.
While the first quality provides perfect ground truth, the second considers measurements with the distance-dependent noise characteristics of the \textit{Azure Kinect} time-of-flight sensor \cite{Toelgyessy2021}.
Smoothed depth is thereby modeled using random Gaussian shifting, and, at very dark surfaces, missing measurements are considered.
Finally, for the \textit{Active Stereo} and \textit{Stereo} qualities, two stereo RGB images with and without a pattern from a simulated dot projector were rendered.
Depth images were then reconstructed using the \textit{\ac{SGM}} algorithm \cite{Hirschmueller2005}.
Examples of the four depth qualities are shown in Fig.\,\ref{fig:e02}.

The benchmark features six robotic systems with different kinematics, ranging from simple open-chain and tree topologies to structures with complex closed kinematics.
Example images of all multi-body objects included in the dataset are shown in Fig.\,\ref{fig:e01}.
An overview of their kinematic structure is given in Tab.\,\ref{tab:e01}. 
\begin{table}
	\caption{
		Characteristic numbers and thresholds for objects\\ included in the \textit{\ac{RTB}} dataset.
	}\label{tab:e01}

\begin{tabularx}{\linewidth}{l | *{6}{>{\centering\arraybackslash}X}}
\hline
\noalign{\smallskip}
&
\rotatebox{90}{Gripper} &
\rotatebox{90}{\parbox{1.15cm}{Medical\\Pliers}} &
\rotatebox{90}{\parbox{1.15cm}{Medical\\Robot}} &
\rotatebox{90}{\parbox{1.15cm}{Picker\\Robot}} &
\rotatebox{90}{\parbox{1.15cm}{Robot\\Fingers}} &
\rotatebox{90}{\parbox{1.15cm}{Robot\\Wrist}}\\
\noalign{\smallskip}
\hline
\noalign{\smallskip}
Number of Bodies & 9 & 5 & 8 & 11 & 16 & 12\\
Degrees of Freedom & 2 & 4 & 7 & 3 & 15 & 2\\
Closed Chains & 4 & 0 & 0 & 6 & 0 & 4\\
Error Threshold $e_\textrm{t}$ [m] & 0.01 & 0.01 & 0.1 & 0.1 & 0.05 & 0.05\\
\noalign{\smallskip}
\hline
\end{tabularx}
\end{table}%
For each robotic system, we provide three difficulty levels: \textit{Easy}, \textit{Medium}, and \textit{Hard}.
In all sequences, the kinematic system is in motion.
However, while for \textit{Easy} sequences, the camera is mostly static with respect to the robot, \textit{Medium} and \textit{Hard} sequences feature faster and shakier motions for both the robot and camera.
Consequently, motion blur increases, which also reduces the quality of stereo matching.
Finally, for each object, difficulty level, and depth image quality, $10$ sequences with $150$ frames were rendered.
In total, this results in $108.000$ frames that feature different kinematic structures, motion patterns, depth measurements, scenes, and lighting conditions.
In summary, given the diverse data, the \textit{\acl{RTB}} allows to extensively measure, compare, and ablate the performance of multi-body tracking algorithms, which is essential for further progress in the field.

\subsection{Metrics}\label{ssec:metrics}
For the evaluation on the \textit{\ac{RTB}} dataset, the \acs{ADD} and \acs{ADD-S} area-under-curve scores \cite{Hinterstoisser2013} are used, which we modify for multi-body structures.
The \acf{ADD} error $e_\textrm{ADD}$ and \acf{ADD-S} error $e_\textrm{ADD-S}$ are thereby computed for each body and frame as follows
\begin{align}\label{eq:e00}
	e_\text{ADD} &= \frac{1}{n_\textrm{v}}\sum_{i=1}^{n_\textrm{v}}\big\lVert \big({}_\text{M}\pmb{\widetilde{X}}_i -  {}_\text{M}\pmb{T}_{\text{M}_\text{GT}}\,{}_\text{M}\pmb{\widetilde{X}}_i\big)_{3\times 1}\big\rVert_2,\\\label{eq:e01}
	e_\text{ADD\textnormal{-}S} &= \frac{1}{n_\textrm{v}}\sum_{i=1}^{n_\textrm{v}} \min_{j\in[n_\textrm{v}]} \big\lVert \big({}_\text{M}\pmb{\widetilde{X}}_i -  {}_\text{M}\pmb{T}_{\text{M}_\text{GT}}\,{}_\text{M}\pmb{\widetilde{X}}_j\big)_{3\times 1}\big\rVert_2,
\end{align}
where ${}_\textrm{M}\pmb{T}_{\textrm{M}_\textrm{GT}}$ is the transformation between the ground truth and estimated model pose, $\pmb{\widetilde{X}}_i$ is a vertex from the 3D mesh of the body written in homogeneous coordinates, $n_\textrm{v}$ is the number of vertices, and $()_{3\times1}$ denotes the first three elements of a vector.
Note that, in the case that a rigid body is composed of multiple 3D meshes, one simply averages the conventional and shortest distance error over all meshes.
Based on the errors of individual bodies and frames $e_{ij}$, we compute the area-under-curve score for an entire kinematic structure and sequence as follows
\begin{equation}\label{eq:e02}
	s = \frac{1}{n_\textrm{b} n_\textrm{f}}\sum_{i=1}^{n_\textrm{b}}\sum_{j=1}^{n_\textrm{f}} \max\Big(1 - \frac{e_{ij}}{e_\textrm{t}}, 0\Big),
\end{equation}
with $n_\textrm{b}$ the number of bodies, $n_\textrm{f}$ the number of frames, and $e_\textrm{t}$ an error threshold.
For the computation of \ac{ADD} and \ac{ADD-S} area-under-curve scores $s_\textrm{ADD}$ and $s_\textrm{ADD-S}$, we use the error metrics $e_\textrm{ADD}$ and $e_\textrm{ADD-S}$, respectively.
Error thresholds $e_\textrm{t}$ that are used for the considered multi-body objects are given in Tab.\,\ref{tab:e01}.
They are defined to be approximately ten percent of each object's diameter.
The resulting score considers each body equally important, takes into account the size of the kinematic structure, and rewards accuracy while limiting the impact of bad predictions.

\subsection{Kinematic Configuration}\label{ssec:e1}
Using the \textit{\ac{RTB}} dataset, we thoroughly evaluate the proposed multi-body tracking framework.
For this, we compare four kinematic configurations per object.
We thereby differentiate between ({\romannumeral 1}) \textit{independently} tracked bodies, ({\romannumeral 2}) \textit{projection} to a minimal parameterization with Jacobians, ({\romannumeral 3}) \textit{constraints} using Lagrange multipliers, and ({\romannumeral 4}) a \textit{combination} of projection and constraints.
In the \textit{combined} case, the main configuration is equal to the \textit{projected} scenario.
However, all chains are closed using additional constraints from the \textit{constrained} version.

Results of the evaluation are shown in Tab.\,\ref{tab:e10}.
\begin{table}
	\caption{
		Comparison of different kinematic configurations on the \textit{\ac{RTB}} dataset,
		showing \ac{ADD-S} area-under-curve scores in percent and average runtimes in milliseconds.
		Tree-like structures are indicated by a $^\star$.
	}\label{tab:e10}

\begin{tabularx}{\linewidth}{l | *{7}{>{\centering\arraybackslash}X} | >{\centering\arraybackslash}X }
	\hline
	\noalign{\smallskip}
	\parbox[b]{1cm}{Config-urations}&
	\rotatebox{90}{\parbox{1.15cm}{Gripper}} &
	\rotatebox{90}{\parbox{1.15cm}{Medical\\Pliers$^\star$}} &
	\rotatebox{90}{\parbox{1.15cm}{Medical\\Robot$^\star$}} &
	\rotatebox{90}{\parbox{1.15cm}{Picker\\Robot}} &
	\rotatebox{90}{\parbox{1.15cm}{Robot\\Fingers$^\star$}} &
	\rotatebox{90}{\parbox{1.15cm}{Robot\\Wrist}}&
	\rotatebox{90}{\textbf{Average}}&
	\rotatebox{90}{\parbox{1.15cm}{\textbf{Runtime}\\ \textbf{[ms]}}}\\
	\noalign{\smallskip}
	\hline
	\noalign{\smallskip}
	Independent & 21.6 & 34.6 & 52.1 & 21.0 & 28.5 & 28.3 & 31.0 & 13.2\\
	Projected & 71.1 & 81.7 & 95.1 & 54.0 & 90.0 & 86.1 & 79.7 & 13.5\\
	Constrained & 87.4 & 79.9 & 92.9 & 93.7 & 84.3 & 96.8 & 89.2 & 16.2\\
	Combined & 87.9 & 81.7 & 95.1 & 94.7 & 90.0 & 97.4 & 91.1 & 13.8\\
	\noalign{\smallskip}
	\hline
\end{tabularx}
\end{table}
The experiments clearly demonstrate that \textit{independent} tracking of bodies is not an option for kinematic structures.
Also, while the \textit{projection} to a minimal parameterization works very well for tree-like structures, for objects with closed kinematic chains, such as the \textit{Gripper}, \textit{Picker Robot}, and \textit{Robot Wrist}, it is a significant disadvantage to not use the full kinematic information.
Finally, what is most interesting is the comparison of the \textit{constrained} and \textit{combined} configurations.
Even though both include the full kinematic information, the \textit{combined} version performs better than the \textit{constrained}.
The main reason for this is a difference in regularization.
For the \textit{constrained} configuration, the regularization considers the full pose variation of each body independently.
On the other hand, for the \textit{combined} case, relative joint variations between bodies are regularized.
Since bodies in a kinematic structure move relative to each other and the entire structure moves with respect to the camera, the relative regularization of the \textit{combined} configuration is closer to reality.
Consequently, using projection and combining it with constraints to model closed chains works best.

\begin{figure}[t]
	\centering

\begin{tikzpicture}
	\small
	\begin{axis}[
		line width=0.7pt,
		width=8.2cm,
		height=5.0cm,
		xlabel={Number of Bodies},
		ylabel={Optimization Time [$ms$]},
		xlabel near ticks,
		ylabel near ticks,
		xmin=0, xmax=50,
		ymin=0, ymax=6,
		xtick={0,10,20,30,40,50},
		ytick={0,2,4,6},
		tick style={draw=none},
		grid=both,
		legend pos=north west,
		legend cell align=left]
		\addplot[color=dlrred] coordinates {	
			(1,0.002697)(2,0.003211)(3,0.004318)(4,0.005421)(5,0.007363)(6,0.009449)(7,0.010308)(8,0.010914)(9,0.013513)(10,0.015317)(11,0.017021)(12,0.018081)(13,0.021017)(14,0.023081)(15,0.025729)(16,0.025065)(17,0.030792)(18,0.032447)(19,0.03567)(20,0.035713)(21,0.041163)(22,0.044516)(23,0.048475)(24,0.04607)(25,0.052297)(26,0.05559)(27,0.060605)(28,0.05978)(29,0.06778)(30,0.071435)(31,0.077286)(32,0.07246)(33,0.081535)(34,0.086169)(35,0.093173)(36,0.090139)(37,0.100682)(38,0.105652)(39,0.113345)(40,0.106209)(41,0.117533)(42,0.124057)(43,0.132375)(44,0.132452)(45,0.145948)(46,0.151985)(47,0.162081)(48,0.151062)(49,0.169066)(50,0.174745)
		};
		\addlegendentry{Projected}
		\addplot[color=dlrblue] coordinates {
			(1,0.008974)(2,0.012762)(3,0.023746)(4,0.026746)(5,0.042664)(6,0.051787)(7,0.071725)(8,0.07904)(9,0.115697)(10,0.126748)(11,0.168482)(12,0.173088)(13,0.237156)(14,0.249874)(15,0.321198)(16,0.318156)(17,0.426125)(18,0.431908)(19,0.560967)(20,0.537109)(21,0.703481)(22,0.696569)(23,0.86808)(24,0.822342)(25,1.07213)(26,1.04732)(27,1.2903)(28,1.21077)(29,1.56301)(30,1.49443)(31,1.85712)(32,1.70301)(33,2.14884)(34,2.06503)(35,2.4971)(36,2.31634)(37,2.89766)(38,2.79998)(39,3.32829)(40,3.10085)(41,3.84116)(42,3.67392)(43,4.33493)(44,4.02401)(45,4.94172)(46,5.25222)(47,6.05024)(48,5.71664)(49,6.80541)(50,6.56847)
		};
		\addlegendentry{Constrained}
		\end{axis}
\end{tikzpicture}
	\caption{
		Optimization time per iteration for the \textit{projected} and \textit{constrained} configuration over the number of bodies in a kinematic chain.
	}\label{fig:e10}
\end{figure}
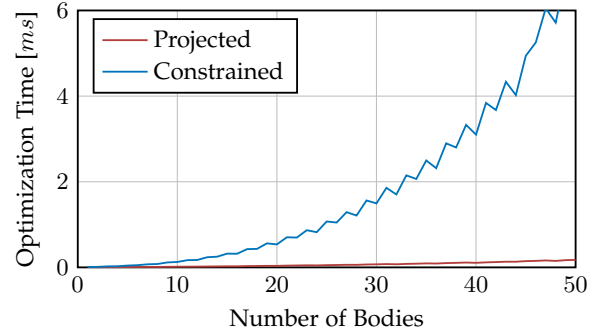

In addition to tracking quality, we also compare computational efficiency.
Results in Tab.\,\ref{tab:e10} show that runtimes for the \textit{independent}, \textit{projected}, and \textit{combined} configurations are almost equal.
For the \textit{constrained} configuration, it is, however, noticeably larger.
To analyze this difference in more detail, we create an additional experiment that compares the \textit{projected} and \textit{constrained} configurations.
We thereby consider a kinematic chain with one rotational link per body and a variable number of bodies.
For the comparison, we measure the runtime required for a single iteration of the optimization and the pose update.
Results over the number of bodies are given in Fig.\,\ref{fig:e10}.
The plots show a significant difference between the two configurations.
The main reason is that, while for the \textit{constrained} version, $11$ unknowns are added per additional body, with $6$ parameters for the body pose and $5$ for the constraint, only $1$ unknown is required in the \textit{projected} scenario.
Together with the Cholesky decomposition's computational complexity of $\mathcal{O}(n^3)$, which is used to solve the linear system of equations, this explains the obtained results.
Also, while the difference between the two formulations decreases for joints with more than one degree of freedom, the \textit{constrained} configuration is never more efficient than the \textit{projected}.
In conclusion, experiments in this section demonstrate that using projection where possible and combining it with constraints where necessary works best both for quality and efficiency.

\subsection{Constraint Convergence}\label{ssec:e2}
\begin{figure}[t]
	\centering
	\input{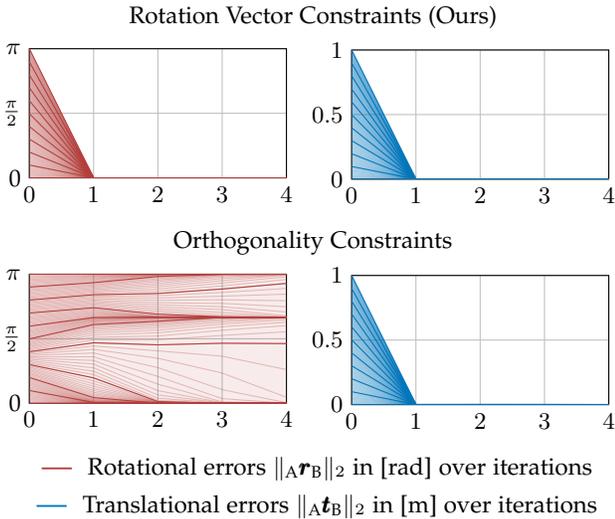}
	\caption{
		Convergence plots showing rotational and translational errors over iterations of the Newton method.
		Initial errors are uniformly distributed with a maximum of $\lVert{}_\textrm{A}\pmb{r}_\textrm{B}\rVert_2 = \pi$ and  $\lVert{}_\textrm{A}\pmb{t}_\textrm{B}\rVert_2 = 1\,\unit{m}$.
		For the translation, the constraint ${}_\textrm{A}\pmb{t}_\textrm{B} = \pmb{0}$ is used.
		For the rotation, we differentiate between our constraints, which restrict the rotation vector using ${}_\textrm{A}\pmb{r}_\textrm{B} = \pmb{0}$, and orthogonality constraints that ensure that axes of reference frame $\textrm{A}$ are orthogonal to axes of reference frame $\textrm{B}$.
		Dark lines visualize error deciles while bright lines show error percentiles.
	}\label{fig:e20}
\end{figure}

In Appendix\,\ref{sec:a3} we proved mathematically that constraints converge in a single iteration for the special case of equal model and constraint coordinate frames.
In the following, we analyze the more general case of unequal coordinate frames.
For this, random transformations ${}_{\textrm{A}}\pmb{T}_{\textrm{M}_a}$ and ${}_{\textrm{B}}\pmb{T}_{\textrm{M}_b}$ are defined.
Also, we start with an initial random difference ${}_\textrm{A}\pmb{T}_\textrm{B}$ between constraint reference frames $\textrm{A}$ and $\textrm{B}$.
Transformations are thereby generated using normalized 3D rotation and translation vectors with a random orientation from a uniform distribution.
The length of those rotation and translation vectors is then sampled from uniform distributions on the intervals $[-1,1]$ and $[-\pi,\pi]$, with translations in meters and rotations in radians.
Given an initial pose difference ${}_\textrm{A}\pmb{T}_\textrm{B}$, four Newton iterations are conducted.
For each iteration, absolute rotation and translation errors $\lVert{}_\textrm{A}\pmb{r}_\textrm{B}\rVert_2$ and $\lVert{}_\textrm{A}\pmb{t}_\textrm{B}\rVert_2$ are computed.
In total, we evaluate $100.000$ random cases and report error percentiles.
Obtained results are visualized in Fig.\,\ref{fig:e20}.
\begin{table*}
	\caption{
		Comparison of tracking algorithms on the \textit{\ac{RTB}} dataset,
		showing \ac{ADD} and \ac{ADD-S} area-under-curve scores in percent\\ and average runtimes in milliseconds.\textsuperscript{\ref{picker_eval}}
		Objects with tree-like kinematic structures are indicated by a $^\star$.
	}\label{tab:e30}

\centering
\begin{tabularx}{\textwidth}{l X *{7}{| >{\centering\arraybackslash}p{0.9cm}@{\hspace{-0.05cm}} >{\centering\arraybackslash}p{0.9cm}}}
\hline
\noalign{\smallskip}
Object&&
\multicolumn{2}{c|}{Gripper}&
\multicolumn{2}{c|}{\!Medical Pliers$^\star$\!\!}&
\multicolumn{2}{c|}{\!Medical Robot$^\star$\!\!}&
\multicolumn{2}{c|}{\!Picker Robot\!}&
\multicolumn{2}{c|}{\!Robot Fingers$^\star$\!\!}&
\multicolumn{2}{c|}{\!Robot Wrist\!}&
\multicolumn{2}{c}{\textbf{Average}}\\
\midrule
App. & Level & ADD & ADD-S & ADD & ADD-S & ADD & ADD-S & ADD & ADD-S & ADD & ADD-S & ADD & ADD-S & ADD & ADD-S \\
\midrule
\multirow{5}{0.95cm}{DART\cite{Schmidt2015}}
& Easy & 33.9 & 57.2 & 29.7 & 44.9 & 75.1 & 86.3 & 27.9 & 33.8 & 54.3 & 64.8 & 51.6 & 63.3 & 45.4 & 58.4\\
& Medium & 1.4 & 3.1 & 11.0 & 25.5 & 12.1 & 20.7 & 3.2 & 5.4 & 2.9 & 4.7 & 4.9 & 8.3 & 5.9 & 11.3\\
& Hard & 1.4 & 3.1 & 8.1 & 20.1 & 3.6 & 7.8 & 1.0 & 2.2 & 1.2 & 2.1 & 1.4 & 2.6 & 2.8 & 6.3\\
& All & 12.2 & 21.2 & 16.2 & 30.1 & 30.2 & 38.3 & 10.7 & 13.8 & 19.5 & 23.9 & 19.3 & 24.7 & 18.0 & 25.3\\
\cmidrule{2-16}
& Runtime & \multicolumn{2}{c|}{4.1\,\unit{ms}} & \multicolumn{2}{c|}{3.7\,\unit{ms}} & \multicolumn{2}{c|}{5.8\,\unit{ms}} & \multicolumn{2}{c|}{9.6\,\unit{ms}} & \multicolumn{2}{c|}{12.8\,\unit{ms}} & \multicolumn{2}{c|}{11.4\,\unit{ms}} & \multicolumn{2}{c}{7.9\,\unit{ms}}\\
\midrule
\multirow{5}{0.95cm}{Mb-ICG\!\!\\ (Ours)}
& Easy & 92.1 & 94.1 & 78.3 & 86.8 & 93.1 & 97.0 & 96.0 & 97.3 & 95.6 & 96.8 & 97.9 & 98.5 & 92.2 & 95.1\\
& Medium & 81.5 & 86.7 & 64.6 & 79.0 & 86.5 & 94.2 & 90.8 & 93.6 & 86.9 & 91.0 & 96.6 & 97.8 & 84.4 & 90.4\\
& Hard & 78.6 & 83.0 & 62.9 & 79.2 & 81.8 & 94.1 & 90.7 & 93.0 & 76.5 & 82.2 & 94.0 & 95.9 & 80.7 & 87.9\\
& All & 84.1 & 87.9 & 68.6 & 81.7 & 87.1 & 95.1 & 92.5 & 94.7 & 86.3 & 90.0 & 96.2 & 97.4 & 85.8 & 91.1\\
\cmidrule{2-16}
& Runtime & \multicolumn{2}{c|}{9.7\,\unit{ms}} & \multicolumn{2}{c|}{6.8\,\unit{ms}} & \multicolumn{2}{c|}{20.5\,\unit{ms}} & \multicolumn{2}{c|}{11.4\,\unit{ms}} & \multicolumn{2}{c|}{16.7\,\unit{ms}} & \multicolumn{2}{c|}{17.5\,\unit{ms}} & \multicolumn{2}{c}{13.8\,\unit{ms}}\\
\hline
\end{tabularx}
\end{table*}%
The plots demonstrate that, like in our proof, both rotational and translational constraints converge in a single iteration.
Also, while we do not report them here, similar results are obtained if individual components of the rotation and translation are constrained.

To provide additional context, we compare our results to convergence plots that use orthogonality constraints for the rotation.
One thereby enforces orthogonality between the axes of the constraint coordinate frames $\textrm{A}$ and $\textrm{B}$ as follows
\begin{equation}\label{eq:e20}
	b_{\textrm{r}ijab}(\pmb{\theta}_a, \pmb{\theta}_b) = {}_\textrm{A}\pmb{e}_i^\top {}_\textrm{A}\pmb{R}_\textrm{B}(\pmb{\theta}_a, \pmb{\theta}_b)\,{}_\textrm{B}\pmb{e}_j,
\end{equation}
with the tuple $(i,j) \in \{(\textrm{x}, \textrm{y}), (\textrm{y}, \textrm{z}), (\textrm{z}, \textrm{x})\}$.
Derivatives that are required for the implementation are given in Appendix\,\ref{sec:a5}.
Convergence plots of the experiment are again shown in Fig.\,\ref{fig:e20}.
The results demonstrate that multiple iterations are required to converge to a kinematically compliant result.
However, what is even worse is that the orthogonality requirement is not unique over the space of possible rotations.
In total, $7$ additional pose configurations exist that fulfill the defined orthogonality constraints.
In Fig.\,\ref{fig:e20}, those cases are visible in the convergence towards a rotational error of $\pi$ and $\frac{2}{3}\pi$.
In conclusion, our experiments demonstrate that, in contrast to other approaches, the developed constraints converge to a kinematically exact solution over the entire space of possible rotations while requiring only a single iteration.

\subsection{Multi-body ICG}\label{ssec:e3}
In addition to the framework's evaluation, we compare our \textit{Multi-body ICG} tracker to \textit{DART} \cite{Schmidt2015, Schmidt2015a}.
\textit{DART} is one of the few general-purpose algorithms for the tracking of kinematic structures that is publicly available.
It uses signed distance functions to consider depth information and allows the configuration of kinematic structures with both rotational and prismatic joints.
For our evaluation, we use the available source code\footnote{\url{https://github.com/tschmidt23/dart}} and leave all parameters at their default values.
Except for the \textit{Picker Robot}, all experiments are conducted on the same computer.%
\footnote{
	\label{picker_eval}Due to memory requirements of the \textit{Picker Robot}, the object's evaluation with \textit{DART} had to be conducted on a computer with an \textit{Intel Xeon Gold 6254} CPU and a \textit{NVIDIA RTX A6000} GPU.
}
Results of the evaluation are shown in Tab.\,\ref{tab:e30}.
The comparison emphasizes the performance of our \textit{Multi-body ICG} tracker.
While for \textit{Easy} sequences, \textit{DART} is still able to track some objects well, for \textit{Medium} and \textit{Hard}, most sequences are impossible.
The main reason is that, due to the large pose differences between frames, the algorithm quickly loses most objects.
This suggests that using depth alone and not being able to model closed kinematic structures significantly limits tracking performance.
In comparison, our \textit{Multi-body ICG} approach is able to successfully track almost all objects and achieves relatively high scores even for \textit{Hard} sequences.

With respect to runtime, both \textit{DART} and our \textit{Multi-body ICG} tracker are able to operate well above $30\,\unit{Hz}$, which is the frequency of most consumer \mbox{RGB-D} cameras.
For \textit{DART}, the largest runtime is observed for the \textit{Robot Fingers}, which has the maximum number of bodies.
For our approach, most resources are required for the tracking of the \textit{Medical Robot}, which uses a large number of correspondence lines and points.
With its highly-optimized \textit{CUDA} implementation and the used \textit{NVIDIA RTX A5000} GPU, \textit{DART} is even faster than our approach.
At the same time, for our method, the \textit{GPU} is only required to validate contour and surface points.
Consequently, it remains mostly idle.
Also, the algorithm runs on a single CPU core.
In summary, while differences with respect to runtime and required computational resources exist, both \textit{DART} and \textit{Multi-body ICG} are highly efficient and provide real-time performance.

Finally, in addition to difficulty levels, we evaluate both approaches with respect to different depth image qualities provided in the \textit{RTB} dataset.
Results of the experiments are given in Tab.\,\ref{tab:e31}.
\begin{table}
	\caption{
		Tracking results on the \textit{\ac{RTB}} dataset for different depth image qualities, with \ac{ADD} and \ac{ADD-S} area-under-curve scores in percent.
	}\label{tab:e31}

\centering
\begin{tabularx}{0.82\linewidth}{X *{2}{| >{\centering\arraybackslash}p{1.0cm}@{\hspace{-0.01cm}} >{\centering\arraybackslash}p{1.0cm}}}
\hline
\noalign{\smallskip}
Approach&
\multicolumn{2}{c|}{DART}&
\multicolumn{2}{c}{Mb-ICG (Ours)}\\
\noalign{\smallskip}
\hline
\noalign{\smallskip}
Depth Quality & ADD & ADD-S & ADD & ADD-S\\
\noalign{\smallskip}
\hline
\noalign{\smallskip}
Ground Truth & 21.1 & 29.0 & 88.7 & 93.0\\
Azure Kinect & 16.2 & 22.4 & 87.9 & 92.5\\
Active Stereo & 18.0 & 25.3 & 86.2 & 91.5\\
Stereo & 16.9 & 24.5 & 80.4 & 87.6\\
\noalign{\smallskip}
\hline
\end{tabularx}
\end{table}%
For \textit{DART}, the evaluation shows a notable difference between \textit{Ground Truth} and other depth image qualities that are more realistic and imperfect.
In comparison, for our approach, differences between \textit{Ground Truth}, \textit{Azure Kinect}, and \textit{Active Stereo} are less severe, and results are mostly comparable.
Only for \textit{Stereo}, for which depth images often have missing measurements and provide considerably less information, the tracking performance is lower.
In summary, the results demonstrate that our \textit{Multi-body ICG} approach does not require perfect depth images.
Instead, its combination of region and depth information ensures high-quality tracking even in challenging scenarios with imperfect depth measurements.
Note that additional experiments that demonstrate the performance for real-world sequences are provided in the supplementary video.

\subsection{Limitations}\label{ssec:e4}
While our framework allows to model various kinematic structures and is compatible with a wide variety of algorithms, some limitations remain.
First, it focuses on approaches that employ Newton-like optimization techniques such as Gauss-Newton, Levenberg-Marquardt, or quasi-Newton methods.
The framework is, therefore, not compatible with particle-based methods or existing deep learning-based techniques.
Also, in this work, we only derived equations based on the axis-angle representation.
Note, however, that it is relatively easy to derive equations required for other parameterizations.
Another limitation comes from the Newton method itself.
Because each iteration considers the energy function only at a single point, the algorithm can get stuck in local minima.
Like for most tracking methods, this limits maximum configuration changes and pose differences between frames.
Also, algorithms that should be extended with our framework have to deal with multiple bodies at the same time.
Typical challenges thereby include bodies that are very close together or even intersect.
Like in our extension of \textit{ICG}, it might be necessary to modify the original algorithm to successfully deal with such cases.

Finally, if our framework is applied to a specific approach, strengths and weaknesses of the original method remain mostly unchanged.
For example, the original \textit{ICG} algorithm requires objects to be distinguishable from the background and needs reasonable depth measurements for the object's surface.
In addition, the algorithm might struggle in situations where the object geometry is not conclusive, such as, for example, in the case of a rotationally symmetric cylinder.
Naturally, the same characteristics also apply to our extension.
Fortunately, strengths of the original algorithm are also preserved.
Like \textit{ICG}, the developed \textit{Multi-body ICG} tracker is therefore highly efficient, accurate, and robust, without requiring object texture.


\section{Conclusion}\label{sec:c}
In this work, we developed a framework that allows to extend existing \ac{6DoF} algorithms to multi-body object tracking.
It combines projection to a minimal parameterization and the application of pose constraints into a single formulation.
With the efficiency and realistic regularization of projection and the ability to model temporary connections and additional constraints, it combines the best of both worlds.
The resulting formulation allows to model a wide variety of kinematic structures and, to the best of our knowledge, is the first that accurately considers closed kinematic chains.
In a detailed mathematical proof, as well as in experiments, we were able to show that the developed constraints enforce an exact kinematic solution and converge in a single iteration of the Newton optimization.
In contrast to previous work, our constraints directly operate on pose differences instead of velocities.
This ensures that kinematic errors are minimized and can not accumulate over time.

Based on the developed framework, we extended the \ac{6DoF} object tracker \textit{ICG} \cite{Stoiber2022} to multi-body tracking.
For a thorough, quantitative evaluation, we created the \textit{\ac{RTB}} dataset.
It features highly-realistic sequences and multiple robots, both with open and closed kinematic chains.
The dataset also provides numerous sequences in various settings with distinct difficulty levels and depth qualities.
In a detailed comparison, we demonstrated that our \textit{Multi-body ICG} approach significantly outperforms \textit{DART} \cite{Schmidt2015}, which is a state-of-the-art articulated object tracker.
With the obtained results and new possibilities for quantitative evaluations, we are confident that both our tracker and dataset will find many applications in robotics and augmented reality.
Also, given the versatility of our framework and its compatibility with a wide variety of existing and potential future methods, we hope that more approaches will move from rigid objects to kinematic structures.

\appendices

\setcounter{equation}{30}

\section{Derivatives of Constraint Equations}\label{sec:a0}
In the following, we derive the first-order derivatives of the extended constraint equation $\pmb{\bar{b}}(\pmb{\theta}_a, \pmb{\theta}_b)$ with respect to $\pmb{\theta}_a$ and $\pmb{\theta}_b$.
We thereby consider the rotational constraint ${}_{\textrm{A}}\pmb{r}_{\textrm{B}}(\pmb{\theta}_a, \pmb{\theta}_b)$ and the translational constraint ${}_{\textrm{A}}\pmb{t}_{\textrm{B}}(\pmb{\theta}_a, \pmb{\theta}_b)$ separately.
Constraint equations are introduced in Section\,\ref{ssec:k2}.

\subsection{Rotational Constraint}\label{sec:a00}
Starting from (\ref{eq:k20}), the variated rotation between the coordinate frames $\textrm{A}$ and $\textrm{B}$ can be expressed using the exponential map as follows
\begin{equation}\label{eq:10}
	\begin{split}
		\exp\big([{}_{\textrm{A}}\pmb{r}_{\textrm{B}}(\pmb{\theta}_{\textrm{r}a}, \pmb{\theta}_{\textrm{r}b})]_\times\big) =\,
		&\exp\big([-{}_{\textrm{A}}\pmb{R}_{\textrm{M}_a}\,\pmb{\theta}_{\textrm{r}a}]_\times\big)\\
		&\exp\big([{}_{\textrm{A}}\pmb{R}_{\textrm{M}_b}\,\pmb{\theta}_{\textrm{r}b}]_\times\big)\\
		&\exp\big([{}_{\textrm{A}}\pmb{r}_{\textrm{B}}]_\times\big),
	\end{split}
\end{equation}
The vectors $\pmb{\theta}_{\textrm{r}a}$ and $\pmb{\theta}_{\textrm{r}b}$ are the rotational components of the pose variation vectors $\pmb{\theta}_{a}$ and $\pmb{\theta}_{b}$.
The constant rotation vector ${}_{\textrm{A}}\pmb{r}_{\textrm{B}}$ is evaluated at $\pmb{\theta}_{\textrm{r}a}=\pmb{\theta}_{\textrm{r}b}=\pmb{0}$.
Note that for the calculation of (\ref{eq:10}), the following relation was used
\begin{equation}\label{eq:11}
	\pmb{R}\exp([\pmb{x}]_\times)\pmb{R}^{-1} = \exp([\pmb{R}\pmb{x}]_\times).
\end{equation}
Knowing that we only need to calculate the derivative with respect to either $\pmb{\theta}_{\textrm{r}a}$ or $\pmb{\theta}_{\textrm{r}b}$ while the other is zero, we reduce (\ref{eq:10}) to take a single variation vector $\pmb{\theta}$ and write
\begin{equation}\label{eq:12}
	\exp\big([{}_{\textrm{A}}\pmb{r}_{\textrm{B}}(\pmb{\theta})]_\times\big) =
	\exp\big([\pmb{\theta}]_\times\big)
	\exp\big([{}_{\textrm{A}}\pmb{r}_{\textrm{B}}]_\times\big).
\end{equation}
For the vector $\pmb{\theta}$, one can then simply substitute one of the rotated variation vectors $\pmb{\theta} = -{}_{\textrm{A}}\pmb{R}_{\textrm{M}_a}\,\pmb{\theta}_{\textrm{r}a}$ or $\pmb{\theta} = {}_{\textrm{A}}\pmb{R}_{\textrm{M}_b}\,\pmb{\theta}_{\textrm{r}b}$.

While the relation in (\ref{eq:12}) looks relatively simple, it is not trivial to calculate the rotation vector ${}_\textrm{A}\pmb{r}_\textrm{B}(\pmb{\theta})$ from $\pmb{\theta}$ and ${}_\textrm{A}\pmb{r}_\textrm{B}$.
Fortunately, a closed-form solution exists that dates back to Olinde Rodrigues.
For this, each rotation vector has to be split into an axis and an angle.
Using the angles $\gamma,\theta,\alpha\in\mathbb{R}$ and the axes $\pmb{n},\pmb{v},\pmb{e}\in\mathbb{R}^3$ with $\lVert\pmb{n}\rVert_2 =\lVert\pmb{v}\rVert_2 = \lVert\pmb{e}\rVert_2 = 1$, we define ${}_\textrm{A}\pmb{r}_\textrm{B}(\pmb{\theta}) = \gamma\pmb{n}$, $\pmb{\theta}=\theta\pmb{v}$, and ${}_\textrm{A}\pmb{r}_\textrm{B} = \alpha\pmb{e}$.
Based on those values, one can then define the following equations
\begin{equation}\label{eq:13}
	\cos\Big(\frac{\gamma}{2}\Big) = \cos\Big(\frac{\theta}{2}\Big)\cos\Big(\frac{\alpha}{2}\Big)
	-\sin\Big(\frac{\theta}{2}\Big)\sin\Big(\frac{\alpha}{2}\Big) \pmb{v}^\top\pmb{e},
\end{equation}
\begin{equation}\label{eq:14}
	\begin{split}
		\sin\Big(\frac{\gamma}{2}\Big)\pmb{n} &
		=\sin\Big(\frac{\theta}{2}\Big)\cos\Big(\frac{\alpha}{2}\Big)\pmb{v} +\cos\Big(\frac{\theta}{2}\Big)\sin\Big(\frac{\alpha}{2}\Big)\pmb{e}\\ &\quad\quad\quad+\sin\Big(\frac{\theta}{2}\Big)\sin\Big(\frac{\alpha}{2}\Big)\pmb{v} \times \pmb{e}.
	\end{split}
\end{equation}
For a detailed derivation and geometrical explanation of this relation, we point interested readers to \cite{Altmann2005}.
To express the required vector ${}_\textrm{A}\pmb{r}_\textrm{B}(\pmb{\theta})$, we define two variables $x = \cos(\frac{\gamma}{2})$ and $\pmb{y} = \sin(\frac{\gamma}{2})\pmb{n}$.
They are the right-hand sides of (\ref{eq:13}) and (\ref{eq:14}).
Based on their reformulation \mbox{$\gamma = 2\cos^{-1}(x)$} and \mbox{$\pmb{n} = \sin(\frac{\gamma}{2})^{-1}\pmb{y}$}, one can write
\begin{equation}\label{eq:15}
	{}_\textrm{A}\pmb{r}_\textrm{B}(\pmb{\theta}) = \gamma\pmb{n} = \frac{2\cos^{-1}(x)}{\sin(\cos^{-1}(x))}\,\pmb{y} = \frac{2\cos^{-1}(x)}{\sqrt{1-x^2}}\,\pmb{y}.
\end{equation}

Finally, starting from this definition, we calculate the first-order derivative with respect to the scalar parameter $\theta$, which is the magnitude of the variation vector $\pmb{\theta}$
\begin{equation}
	\begin{split}
		\frac{\partial{}_\textrm{A}\pmb{r}_\textrm{B}}{\partial\theta}
		& = \frac{\frac{-2}{\sqrt{1-x^2}}\sqrt{1-x^2} - 2\cos^{-1}(x)\frac{-x}{\sqrt{1-x^2}}}{1-x^2}\frac{\partial x}{\partial\theta}\pmb{y}\\
		&\quad\quad\quad+ \frac{2\cos^{-1}(x)}{\sqrt{1-x^2}}\frac{\partial\pmb{y}}{\partial{\theta}},
	\end{split}
\end{equation}	
Also, based on the right-hand sides of (\ref{eq:13}) and (\ref{eq:14}), we calculate
\begin{align}
	\phantom{\frac{\partial x}{\partial\theta}}
	&\mathllap{\frac{\partial x}{\partial\theta}}
	= -\frac{\sin(\frac{\theta}{2})\cos(\frac{\alpha}{2})}{2} - \frac{\cos(\frac{\theta}{2})\sin(\frac{\alpha}{2})}{2}\pmb{v}^\top\pmb{e},\\[5pt]
	&\begin{aligned}
		\mathllap{\frac{\partial\pmb{y}}{\partial{\theta}}} 	&=\frac{\cos(\frac{\theta}{2})\cos(\frac{\alpha}{2})}{2}\pmb{v} -\frac{\sin(\frac{\theta}{2})\sin(\frac{\alpha}{2})}{2}\pmb{e}\\ &\quad\quad\quad+\frac{\cos(\frac{\theta}{2})\sin(\frac{\alpha}{2})}{2}\pmb{v}\times\pmb{e}.
	\end{aligned}
\end{align}
By evaluating the derivatives at $\theta = 0$, for which, according to (\ref{eq:13}) and (\ref{eq:14}), $x = \cos(\frac{\alpha}{2})$ and $\pmb{y} = \sin(\frac{\alpha}{2})\pmb{e}$, the following equations are obtained
\begin{align}\label{eq:16}
	\frac{\partial{}_\textrm{A}\pmb{r}_\textrm{B}}{\partial\theta}\bigg\vert_{\theta=0} =\,
	&\frac{-2+\alpha\cot(\frac{\alpha}{2})}{\sin(\frac{\alpha}{2})}\pmb{e} \frac{\partial x}{\partial\theta} + \frac{\alpha}{\sin(\frac{\alpha}{2})} \frac{\partial\pmb{y}}{\partial{\theta}} \bigg\vert_{\theta=0},\\[5pt]\label{eq:17}
	\frac{\partial x}{\partial\theta}\bigg\vert_{\theta=0} =\,
	&-\frac{1}{2}\sin\Big(\frac{\alpha}{2}\Big)\pmb{v}^\top\pmb{e},\\[5pt]\label{eq:18}
	\frac{\partial\pmb{y}}{\partial\theta}\bigg\vert_{\theta=0} =\,
	&\frac{1}{2}\cos\Big(\frac{\alpha}{2}\Big)\pmb{v} + \frac{1}{2}\sin\Big(\frac{\alpha}{2}\Big)\pmb{v}\times\pmb{e}.
\end{align}
Introducing (\ref{eq:17}) and (\ref{eq:18}) in (\ref{eq:16}) and extracting the pose variation's rotation axis $\pmb{v}$ results in the final expression for the first-order derivative
\begin{align}\label{eq:19}
	\phantom{\frac{\partial{}_\textrm{A}\pmb{r}_\textrm{B}}{\partial\theta}\bigg\vert_{\theta=0}}
	&\begin{aligned}
		\mathllap{\frac{\partial{}_\textrm{A}\pmb{r}_\textrm{B}}{\partial\theta}\bigg\vert_{\theta=0}}
		&=\Big(1 - \frac{\alpha}{2}\cot\Big(\frac{\alpha}{2}\Big)\Big)\pmb{e}\pmb{v}^\top\pmb{e}\\
		&\quad\quad\quad+\frac{\alpha}{2}\cot\Big(\frac{\alpha}{2}\Big)\pmb{v} +\frac{\alpha}{2}\pmb{v}\times\pmb{e}
	\end{aligned}\\[5pt]\label{eq:20}
	&\begin{aligned}
		&=\bigg(\frac{\alpha}{2}\cot\Big(\frac{\alpha}{2}\Big) \pmb{I} - 	\frac{\alpha}{2}[\pmb{e}]_\times\\
		&\quad\quad\quad+ \Big(1 - 	\frac{\alpha}{2}\cot\Big(\frac{\alpha}{2}\Big)\Big)\pmb{e}\pmb{e}^\top\bigg)\pmb{v}.
	\end{aligned}
\end{align}

Equation\,(\ref{eq:20}) shows that the derivative with respect to the rotation angle $\theta$ consists of a linear combination of column vectors with the rotation axis $\pmb{v}$.
For the rotation vector $\pmb{\theta} = \theta\pmb{v}$, the rotation axis $\pmb{v}$ projects the scalar angle $\theta$ in a similar fashion.
Knowing that
\begin{equation}
\frac{\partial{}_\textrm{A}\pmb{r}_\textrm{B}}{\partial\theta} = \frac{\partial{}_\textrm{A}\pmb{r}_\textrm{B}}{\partial\pmb{\theta}} \frac{\partial\pmb{\theta}}{\partial\theta} = \frac{\partial{}_\textrm{A}\pmb{r}_\textrm{B}}{\partial\pmb{\theta}}\,\pmb{v},
\end{equation}
we are finally able to extract the first-order derivative with respect to the rotation vector $\pmb{\theta}$ from (\ref{eq:20}) and write
\begin{equation}\label{eq:21}
	\begin{aligned}
		\frac{\partial{}_\textrm{A}\pmb{r}_\textrm{B}}{\partial\pmb{\theta}}\bigg\vert_{\pmb{\theta}=\pmb{0}}& = \frac{\alpha}{2}\cot\Big(\frac{\alpha}{2}\Big) \pmb{I} - \frac{\alpha}{2}[\pmb{e}]_\times\\
		&\quad\quad\quad+ \Big(1 - \frac{\alpha}{2}\cot\Big(\frac{\alpha}{2}\Big)\Big)\pmb{e}\pmb{e}^\top.
	\end{aligned}
\end{equation}

For the originally required derivatives of the rotational constraint ${}_\textrm{A}\pmb{r}_\textrm{B}(\pmb{\theta}_a,\pmb{\theta}_b)$ with respect to the rotational pose variation vectors $\pmb{\theta}_{\textrm{r}a}$ and $\pmb{\theta}_{\textrm{r}b}$, one simply substitutes $\pmb{\theta} = -{}_{\textrm{A}}\pmb{R}_{\textrm{M}_a}\,\pmb{\theta}_{\textrm{r}a}$ and $\pmb{\theta} = {}_{\textrm{A}}\pmb{R}_{\textrm{M}_b}\,\pmb{\theta}_{\textrm{r}b}$.
Considering the partial derivatives $\frac{\partial\pmb{\theta}}{\partial\pmb{\theta}_{\textrm{r}a}} = -{}_{\textrm{A}}\pmb{R}_{\textrm{M}_a}$ and $\frac{\partial\pmb{\theta}}{\partial\pmb{\theta}_{\textrm{r}b}} = {}_{\textrm{A}}\pmb{R}_{\textrm{M}_b}$, we then obtain the following final results for the first-order derivatives of the rotational constraint
\begin{align}\label{eq:22}
	\frac{\partial{}_\textrm{A}\pmb{r}_\textrm{B}}{\partial\pmb{\theta}_a}\bigg\vert_{\pmb{\theta}_\textrm{k}=\pmb{0}} =\,&
	\begin{bmatrix}
		-\pmb{C}\,{}_{\textrm{A}}\pmb{R}_{\textrm{M}_a}&\pmb{0}
	\end{bmatrix},\\[5pt]\label{eq:23}
	\frac{\partial{}_\textrm{A}\pmb{r}_\textrm{B}}{\partial\pmb{\theta}_b}\bigg\vert_{\pmb{\theta}_\textrm{k}=\pmb{0}} =\,&
	\begin{bmatrix}
		\pmb{C}\,{}_{\textrm{A}}\pmb{R}_{\textrm{M}_b}&\pmb{0}
	\end{bmatrix}.
\end{align}
The matrix $\pmb{C}$, which we call variation matrix, is thereby defined as
\begin{equation}\label{eq:24}
	\pmb{C} = \frac{\alpha}{2}\cot\Big(\frac{\alpha}{2}\Big) \pmb{I} - \frac{\alpha}{2}[\pmb{e}]_\times + \Big(1 - \frac{\alpha}{2}\cot\Big(\frac{\alpha}{2}\Big)\Big)\pmb{e}\pmb{e}^\top.
\end{equation}
It is the partial derivative of $\frac{\partial{}_\textrm{A}\pmb{r}_\textrm{B}}{\partial\pmb{\theta}}\big\vert_{\pmb{\theta}=\pmb{0}}$ and is computed
using the angle and axis of the current rotation vector ${}_\textrm{A}\pmb{r}_\textrm{B} = \alpha\pmb{e}$.
The variation matrix $\pmb{C}$ describes how the rotation vector ${}_\textrm{A}\pmb{r}_\textrm{B}$ changes for the variation with a subsequent infinitesimal rotation.
Properties of the matrix are derived in Appendix\,\ref{sec:a1}.

\subsection{Translational Constraint}\label{sec:a01}
For the translational constraint, we again start from (\ref{eq:k20})
to formulate the variated translation between the coordinate frames $\textrm{A}$ and $\textrm{B}$.
Using the knowledge that derivatives are either calculated with respect to $\pmb{\theta}_\textrm{a}$ or $\pmb{\theta}_\textrm{b}$, one can write
\begin{align}\label{eq:30}
	{}_{\textrm{A}}\pmb{t}_{\textrm{B}}(\pmb{\theta}_a)
	&={}_\textrm{A}\pmb{R}_{\textrm{M}_a}
		\big(\pmb{R}(-\pmb{\theta}_{\textrm{r}a})
		({}_{\textrm{M}_a}\pmb{t}_{\textrm{B}} - \pmb{\theta}_{\textrm{t}a})\big)
		+ {}_\textrm{A}\pmb{t}_{\textrm{M}_a},\\[5pt]\label{eq:30_1}
	{}_{\textrm{A}}\pmb{t}_{\textrm{B}}(\pmb{\theta}_b)
	&= {}_\textrm{A}\pmb{R}_{\textrm{M}_b}
	\big(\pmb{R}(\pmb{\theta}_{\textrm{r}b})\,
	{}_{\textrm{M}_b}\pmb{t}_{\textrm{B}} + \pmb{\theta}_{\textrm{t}b}\big)
	+ {}_\textrm{A}\pmb{t}_{\textrm{M}_b},
\end{align}
where the vectors $\pmb{\theta}_{\textrm{t}a}$ and $\pmb{\theta}_{\textrm{t}b}$ are the translational components of the variation vectors $\pmb{\theta}_{a}$ and $\pmb{\theta}_{b}$.

Using the Taylor series expansion of the exponential map and the knowledge that $[\pmb{x}]_\times\pmb{y} = -[\pmb{y}]_\times\pmb{x}$, we define the following general relation
\begin{align}\label{eq:31}
	\pmb{R}(\pmb{\theta})\pmb{t}
	&= \exp(\pmb{\theta})\pmb{t}\\[5pt]\label{eq:31a}
	&= (\pmb{I} + [\pmb{\theta}]_\times + \mathcal{O}(\pmb{\theta}^2))\pmb{t}\\[5pt]\label{eq:31b}
	&= \pmb{t} - [\pmb{t}]_\times \pmb{\theta} + \mathcal{O}(\pmb{\theta}^2)\pmb{t},
\end{align}
where the expression $\mathcal{O}(\pmb{\theta}^2)$ models higher-order terms.
Using this relation in (\ref{eq:30}) and (\ref{eq:30_1}) and neglecting higher-order terms, it is straightforward to calculate the final results for the first-order derivatives of the translational constraint
\begin{align}\label{eq:32}
	\frac{\partial{}_\textrm{A}\pmb{t}_\textrm{B}}{\partial\pmb{\theta}_a}\bigg\vert_{\pmb{\theta}_\textrm{k}=\pmb{0}} =\,&
	\begin{bmatrix}
		{}_{\textrm{A}}\pmb{R}_{\textrm{M}_a}\,[{}_{\textrm{M}_a}\pmb{t}_{\textrm{B}}]_\times& -{}_{\textrm{A}}\pmb{R}_{\textrm{M}_a}
	\end{bmatrix},\\[5pt]\label{eq:33}
	\frac{\partial{}_\textrm{A}\pmb{t}_\textrm{B}}{\partial\pmb{\theta}_b}\bigg\vert_{\pmb{\theta}_\textrm{k}=\pmb{0}} =\,&
	\begin{bmatrix}
		-{}_{\textrm{A}}\pmb{R}_{\textrm{M}_b}\,[{}_{\textrm{M}_b}\pmb{t}_{\textrm{B}}]_\times& {}_{\textrm{A}}\pmb{R}_{\textrm{M}_b}
	\end{bmatrix}.
\end{align}

\section{Properties of the Variation Matrix}\label{sec:a1}
The variation matrix $\pmb{C}$, which was derived in Appendix\,\ref{sec:a00} describes how the rotation vector ${}_\textrm{A}\pmb{r}_\textrm{B}$ changes with the variation of a \textit{subsequent} infinitesimal rotation.
It is computed according to (\ref{eq:24}) and depends on the rotational difference between the coordinate frames $\textrm{A}$ and $\textrm{B}$.
In the following, we investigate its relation to the rotation matrix ${}_\textrm{A}\pmb{R}_\textrm{B}$ and the rotation vector ${}_\textrm{A}\pmb{r}_\textrm{B}$.
For brevity, we drop the subscripts $\textrm{A}$ and $\textrm{B}$ in the notation and simply write $\pmb{R}$ and $\pmb{r}$.

Given the rotation vector $\pmb{r} = \alpha\pmb{e}$ with the angle $\alpha$ and the normalized axis $\pmb{e}$, the rotation matrix can be computed using the Rodrigues formula
\begin{equation}\label{eq:a100}
	\pmb{R} = \pmb{I} + \sin(\alpha)[\pmb{e}]_\times + 2 \sin\Big(\frac{\alpha}{2}\Big)^{\hspace{-1pt}2} [\pmb{e}]_\times^2.
\end{equation}
With the knowledge that $[\pmb{e}]_\times\pmb{e} = \pmb{0}$, we are able to calculate the following product for the rotation and variation matrix
\begin{equation}\label{eq:a101}
	\begin{split}
		\pmb{R}\pmb{C} =\,
		&\frac{\alpha}{2}\cot\Big(\frac{\alpha}{2}\Big) \pmb{I} - \frac{\alpha}{2}[\pmb{e}]_\times + \Big(1 - \frac{\alpha}{2}\cot\Big(\frac{\alpha}{2}\Big)\Big)\pmb{e}\pmb{e}^\top\\
		&\quad+ \frac{\alpha}{2}\sin(\alpha)\cot\Big(\frac{\alpha}{2}\Big)[\pmb{e}]_\times - \frac{\alpha}{2}\sin(\alpha)[\pmb{e}]_\times^2\\
		&\quad +\alpha\sin\Big(\frac{\alpha}{2}\Big)^{\hspace{-1pt}2}\cot\Big(\frac{\alpha}{2}\Big)[\pmb{e}]_\times^2 - \alpha\sin\Big(\frac{\alpha}{2}\Big)^{\hspace{-1pt}2}[\pmb{e}]_\times^3.
	\end{split}
\end{equation}
Using the relations
\begin{align}\label{eq:a102}
	\frac{\alpha}{2}\sin(\alpha)\cot\Big(\frac{\alpha}{2}\Big) &= \alpha - \alpha\sin\Big(\frac{\alpha}{2}\Big)^{\hspace{-1pt}2},\\[5pt]\label{eq:a103}
	\alpha\sin\Big(\frac{\alpha}{2}\Big)^{\hspace{-1pt}2}\cot\Big(\frac{\alpha}{2}\Big) &= \frac{\alpha}{2}\sin(\alpha),\\[5pt]\label{eq:a104}
	[\pmb{e}]_\times^3 &= -[\pmb{e}]_\times,
\end{align}
one can simplify (\ref{eq:a101}) and write
\begin{align}\label{eq:a105}
	\pmb{R}\pmb{C} &= \frac{\alpha}{2}\cot\Big(\frac{\alpha}{2}\Big) \pmb{I} + \frac{\alpha}{2}[\pmb{e}]_\times + \Big(1 - \frac{\alpha}{2}\cot\Big(\frac{\alpha}{2}\Big)\Big)\pmb{e}\pmb{e}^\top\\[5pt]\label{eq:a106}
	&=\pmb{C}^\top,
\end{align}
which shows that the product of the rotation and variation matrix is simply the transposed variation matrix.
Looking at (\ref{eq:13}) and (\ref{eq:14}), one notices that changing the order of the rotations $\pmb{r}$ and $\pmb{\theta}$ only changes the sign in the cross product term in (\ref{eq:14}).
Deriving the corresponding variation matrix then leads to the same expression as in (\ref{eq:a105}), which is simply the transposed of the original variation matrix $\pmb{C}$.
The transposed variation matrix $\pmb{C}^\top$ therefore describes how the rotation vector $\pmb{r}$ changes with the variation of a \textit{preceding} infinitesimal rotation.
Based on this interpretation, (\ref{eq:a106}) shows that rotating the variation matrix $\pmb{C}$ with the rotation matrix $\pmb{R}$ changes the order in which the infinitesimal rotation $\pmb{\theta}$ is applied to the rotation $\pmb{r}$.

Similarly, transposing the relation in (\ref{eq:a106}) and using the knowledge that for valid rotation matrices $\pmb{R}^\top = \pmb{R}^{-1}$, one is able to write
\begin{equation}\label{eq:a107}
	\pmb{R}\pmb{C} = \pmb{C}^\top \;\Leftrightarrow\; \pmb{C}^\top\pmb{R}^\top = \pmb{C} \;\Leftrightarrow\; \pmb{C}^\top = \pmb{C}\pmb{R}.
\end{equation}
This shows that $\pmb{R}$ and $\pmb{C}$ are commutative. 
Hence, applying the rotation matrix $\pmb{R}$ on either side of the variation matrix $\pmb{C}$ changes the location of the infinitesimal rotation from a \textit{subsequent} to a \textit{preceding} rotation.
Finally, based on the relations in (\ref{eq:a107}), we are able to express the rotation matrix with respect to the variation matrix as follows
\begin{equation}\label{eq:a108}
	\pmb{R} = \pmb{C}^\top\pmb{C}^{-1} = \pmb{C}^{-1}\pmb{C}^\top.
\end{equation}
The equation shows that because $\pmb{C}$ and $\pmb{C}^\top$ consider the variation with respect to the frames before and after the rotation, the matrix $\pmb{R}$ is implicitly encoded in the variation matrix $\pmb{C}$.
Also, (\ref{eq:a108}) shows that the variation matrix is normal with $\pmb{C}\pmb{C}^\top = \pmb{C}^\top\pmb{C}$.
However, at the same time, experiments show that, with $\pmb{C}^\top \neq \pmb{C}$ and $\pmb{C}^\top \neq \pmb{C}^{-1} $, it is neither symmetric nor orthogonal.

Finally, we also want to analyze how the rotation vector $\pmb{r}$ is connected to the variation matrix $\pmb{C}$.
For this, the product $\pmb{C}\pmb{r}$ is computed.
With the relations $[\pmb{e}]_\times\pmb{e} = \pmb{0}$ and $\pmb{e}^\top\pmb{e} = 1$, we calculate
\begin{align}\label{eq:a109}
	\pmb{C}\pmb{r} &= \alpha\Big(\frac{\alpha}{2}\cot\Big(\frac{\alpha}{2}\Big)\Big) \pmb{e} + \alpha\Big(1 - \frac{\alpha}{2}\cot\Big(\frac{\alpha}{2}\Big)\Big)\pmb{e},\\[5pt]\label{eq:a110}
	&=\pmb{r}.
\end{align}
Given the knowledge that the rotation vector $\pmb{r}$ is the eigenvector of the rotation matrix $\pmb{R}$ for the eigenvalue $\lambda = 1$, one is able to define the relation $\pmb{r} = \pmb{R}\pmb{r}$.
Together with the results from (\ref{eq:a106}), we can therefore show that
\begin{equation}\label{eq:a111}
	\pmb{C}\pmb{r} = \pmb{r} \;\Leftrightarrow\; \pmb{C}\pmb{R}\pmb{r} = \pmb{r} \;\Leftrightarrow\; \pmb{C}^\top\pmb{r} = \pmb{r}.
\end{equation}
This demonstrates that the rotation vector $\pmb{r}$ is an eigenvector of both the original and transposed variation matrix $\pmb{C}$.
Hence, it is an eigenvector of both the variation matrix for a \textit{subsequent} and \textit{preceding} infinitesimal rotation.

\section{Constraint Convergence}\label{sec:a3}
Given the developed constraint equations and first-order derivatives, in the following, we analyze how fast the Newton method converges to a compliant kinematic result.
For this, two bodies with an initial rotational or translational difference of ${}_\textrm{A}\pmb{R}_\textrm{B}$ or ${}_\textrm{A}\pmb{t}_\textrm{B}$ are considered.
Based on this initial error, the relative pose update for a full rotational or translational constraint is analyzed.
Experiments, which demonstrate that the obtained mathematical results also hold for more general cases, are provided in Section\,\ref{ssec:e2}.

\subsection{Rotational Constraint}\label{sec:a30}
Without loss of generality, we assume that constraint coordinate frames are equal to model frames with $\textrm{A} = \textrm{M}_a$ and $\textrm{B} = \textrm{M}_b$.
Also, the translation of both bodies is fixed and the combined variation $\pmb{\theta}_\textrm{k}^\top = \begin{bmatrix} \pmb{\theta}_{\textrm{r}a}^\top & \pmb{\theta}_{\textrm{r}b}^\top \end{bmatrix}$ only considers rotation.
Based on those assumptions, the constraint equation and constraint Jacobian can be written as
\begin{align}\label{eq:a300}
	\pmb{b}_\textrm{k} &= {}_\textrm{A}\pmb{r}_\textrm{B},\\[5pt]\label{eq:a301}
	\pmb{B}_\textrm{k} &= \begin{bmatrix}-\pmb{C} & \pmb{C}\,{}_\textrm{A}\pmb{R}_\textrm{B} \end{bmatrix}.
\end{align}
Introducing both statements in (\ref{eq:m32}) and given some arbitrary Hessian matrices and gradient vectors for body $a$ and $b$, we obtain the linear equation for a single Newton step
\begin{equation}\label{eq:a302}
	\begin{bmatrix}
		-\pmb{H}_a & \pmb{0} & \pmb{C}^\top\\
		\pmb{0} & -\pmb{H}_b & -{}_\textrm{A}\pmb{R}_\textrm{B}^\top\,\pmb{C}^\top\\
		\pmb{C} & -\pmb{C}\,{}_\textrm{A}\pmb{R}_\textrm{B} & \pmb{0}
	\end{bmatrix}
	\begin{bmatrix}
		\pmb{\theta}_{\textrm{r}a}\\
		\pmb{\theta}_{\textrm{r}b}\\
		\pmb{\lambda}
	\end{bmatrix}
	=
	\begin{bmatrix}
		\pmb{g}_a\\
		\pmb{g}_b\\
		{}_\textrm{A}\pmb{r}_\textrm{B}
	\end{bmatrix}.
\end{equation}
Based on the first two lines of the equation, one can directly write 
\begin{align}\label{eq:a304}
	\pmb{\theta}_{\textrm{r}a} &= \pmb{H}_a^{-1}(\pmb{C}^\top\pmb{\lambda} - \pmb{g}_a),\\[5pt]\label{eq:a305}
	\pmb{\theta}_{\textrm{r}b} &= \pmb{H}_b^{-1}(-{}_\textrm{A}\pmb{R}_\textrm{B}^\top\,\pmb{C}^\top\pmb{\lambda} - \pmb{g}_b),
\end{align}
where the only unknown is $\pmb{\lambda}$.
Subsequently, introducing (\ref{eq:a304}) and (\ref{eq:a305}) into the last line of (\ref{eq:a302}) leads to the following expression
\begin{equation}\label{eq:a306}
	\pmb{C}(\pmb{X} + \pmb{Y})\pmb{C}^\top\pmb{\lambda} + \pmb{C}(-\pmb{x} + \pmb{y}) = {}_\textrm{A}\pmb{r}_\textrm{B},
\end{equation}
where the matrices $\pmb{X} = \pmb{H}_a^{-1}$ and $\pmb{Y} = {}_\textrm{A}\pmb{R}_\textrm{B}\,\pmb{H}_b^{-1}{}_\textrm{A}\pmb{R}_\textrm{B}^\top$, as well as the vectors $\pmb{x} = \pmb{H}_a^{-1}\pmb{g}_a$ and $\pmb{y} = {}_\textrm{A}\pmb{R}_\textrm{B}\,\pmb{H}_b^{-1}\pmb{g}_b$ are used for conciseness.
Rewriting this equation leads to an expression for $\pmb{\lambda}$ that only depends on gradient vectors, Hessian matrices, and the current rotational difference
\begin{equation}\label{eq:a307}
	\pmb{\lambda} = \pmb{C}^{-\top}(\pmb{X} + \pmb{Y})^{-1}(\pmb{x} - \pmb{y} + \pmb{C}^{-1} {}_\textrm{A}\pmb{r}_\textrm{B}).
\end{equation}
Finally, using the proof from Appendix~\ref{sec:a1}, which shows that ${}_\textrm{A}\pmb{r}_\textrm{B}$ is an eigenvector of the variation matrix $\pmb{C}$ and that $\pmb{C}^{-1} {}_\textrm{A}\pmb{r}_\textrm{B} = {}_\textrm{A}\pmb{r}_\textrm{B}$, we can further simplify and write
\begin{equation}\label{eq:a308}
	\pmb{\lambda} = \pmb{C}^{-\top}(\pmb{X} + \pmb{Y})^{-1}(\pmb{x} - \pmb{y} + {}_\textrm{A}\pmb{r}_\textrm{B}).
\end{equation}

To assess the convergence of our method, we compute how the rotational difference changes for a single Newton step.
Using the axis-angle representation, the change in rotation is thereby calculated as
\begin{equation}\label{eq:a309}
	\Delta\pmb{R} =
	\exp\big([-\pmb{\theta}_{\textrm{r}a}]_\times\big) \exp\big([{}_{\textrm{A}}\pmb{R}_\textrm{B}\,\pmb{\theta}_{\textrm{r}b}]_\times\big).
\end{equation}
By introducing (\ref{eq:a304}) and (\ref{eq:a305}), one can then write
\begin{equation}\label{eq:a310}
		\Delta\pmb{R} =\exp\big([-\pmb{X}\pmb{C}^\top\pmb{\lambda} + \pmb{x}]_\times\big)
		\exp\big([-\pmb{Y}\pmb{C}^\top\pmb{\lambda} - \pmb{y}]_\times\big).
\end{equation}
Extending this equation with $\pmb{Y} - \pmb{Y}$ and $\pmb{X} - \pmb{X}$ leads to
\begin{equation}\label{eq:a311}
	\begin{split}
		\Delta\pmb{R}&=\exp\Big(\mbox{\large$\frac{1}{2}$}\big[(\pmb{Y}-\pmb{X}-\pmb{X}-\pmb{Y})\pmb{C}^\top\pmb{\lambda} + 2\pmb{x}\big]_\times\Big)\\
		&\quad\quad\exp\Big(\mbox{\large$\frac{1}{2}$}\big[(\pmb{X}-\pmb{Y}-\pmb{X}-\pmb{Y})\pmb{C}^\top\pmb{\lambda} - 2\pmb{y}\big]_\times\Big).
	\end{split}
\end{equation}
Subsequently, we introduce $\pmb{\lambda}$ from (\ref{eq:a308}) and multiply it with the expression $(-\pmb{X} - \pmb{Y})\pmb{C}^\top$ in (\ref{eq:a311}) to obtain
\begin{equation}\label{eq:a312}
	\begin{split}
		\Delta\pmb{R} &=\exp\Big(\mbox{\large$\frac{1}{2}$}\big[(\pmb{Y}-\pmb{X})\pmb{C}^\top\pmb{\lambda} + \pmb{x} + \pmb{y} - {}_\textrm{A}\pmb{r}_\textrm{B}\big]_\times\Big)\\
		&\quad\quad\exp\Big(\mbox{\large$\frac{1}{2}$}\big[(\pmb{X}-\pmb{Y})\pmb{C}^\top\pmb{\lambda} - \pmb{x} - \pmb{y} - {}_\textrm{A}\pmb{r}_\textrm{B}\big]_\times\Big),
	\end{split}
\end{equation}
which has the following highly symmetric structure
\begin{equation}\label{eq:a313}
	\Delta\pmb{R}
	=\exp\Big(\mbox{\large$\frac{1}{2}$}[\pmb{z} - {}_\textrm{A}\pmb{r}_\textrm{B}]_\times\Big)\exp\Big(\mbox{\large$\frac{1}{2}$}[-\pmb{z} - {}_\textrm{A}\pmb{r}_\textrm{B}]_\times\Big).
\end{equation}
Finally, observing that the two exponents are commutative and knowing that commutative exponents can be summed, we get the following result for the rotational difference
\begin{equation}\label{eq:a314}
	\Delta\pmb{R}=\exp\big([-{}_\textrm{A}\pmb{r}_\textrm{B}]_\times\big)
	={}_\textrm{A}\pmb{R}_\textrm{B}^{-1}
\end{equation} 
This shows that a single iteration of the Newton method is enough to minimize an initial rotational error of ${}_\textrm{A}\pmb{R}_\textrm{B}$ to zero and obtain an exact kinematic solution.

\subsection{Translational Constraint}\label{sec:a40}
Without loss of generality, we again assume that constraint coordinate frames are equal to model frames with $\textrm{A} = \textrm{M}_a$ and $\textrm{B} = \textrm{M}_b$.
Furthermore, the rotation of both bodies is fixed and the combined variation vector $\pmb{\theta}_\textrm{k}^\top = \begin{bmatrix} \pmb{\theta}_{\textrm{t}a}^\top & \pmb{\theta}_{\textrm{t}b}^\top \end{bmatrix}$ only considers translation.
The constraint equation and constraint Jacobian can then be written as
\begin{align}\label{eq:a400}
	\pmb{b}_\textrm{k} &= {}_\textrm{A}\pmb{t}_\textrm{B},\\[5pt]\label{eq:a401}
	\pmb{B}_\textrm{k} &= \begin{bmatrix}-\pmb{I} & {}_\textrm{A}\pmb{R}_\textrm{B} \end{bmatrix}.
\end{align}
Introducing both statements in (\ref{eq:m32}), we obtain the following linear equation for a single Newton step
\begin{equation}\label{eq:a402}
	\begin{bmatrix}
		-\pmb{H}_a & \pmb{0} & \pmb{I}^\top\\
		\pmb{0} & -\pmb{H}_b & -{}_\textrm{A}\pmb{R}_\textrm{B}^\top\\
		\pmb{I} & -{}_\textrm{A}\pmb{R}_\textrm{B} & \pmb{0}
	\end{bmatrix}
	\begin{bmatrix}
		\pmb{\theta}_{\textrm{t}a}\\
		\pmb{\theta}_{\textrm{t}b}\\
		\pmb{\lambda}
	\end{bmatrix}
	=
	\begin{bmatrix}
		\pmb{g}_a\\
		\pmb{g}_b\\
		{}_\textrm{A}\pmb{t}_\textrm{B}
	\end{bmatrix}.
\end{equation}
Based on the first two rows of this relation, variation vectors are computed as
\begin{align}\label{eq:a404}
	\pmb{\theta}_{\textrm{t}a} &= \pmb{H}_a^{-1}(\pmb{\lambda} - \pmb{g}_a),\\[5pt]\label{eq:a405}
	\pmb{\theta}_{\textrm{t}b} &= \pmb{H}_b^{-1}(-{}_\textrm{A}\pmb{R}_\textrm{B}^\top\,\pmb{\lambda} - \pmb{g}_b).
\end{align}
For conciseness, we again define the matrices $\pmb{X} = \pmb{H}_a^{-1}$ and $\pmb{Y} = {}_\textrm{A}\pmb{R}_\textrm{B}\,\pmb{H}_b^{-1}{}_\textrm{A}\pmb{R}_\textrm{B}^\top$, as well as the vectors $\pmb{x} = \pmb{H}_a^{-1}\pmb{g}_a$ and $\pmb{y} = {}_\textrm{A}\pmb{R}_\textrm{B}\,\pmb{H}_b^{-1}\pmb{g}_b$.
We then introduce (\ref{eq:a404}) and (\ref{eq:a405}) into the last row of (\ref{eq:a402}) to compute the following relation for $\pmb{\lambda}$ that includes only known variables
\begin{equation}\label{eq:a408}
	\pmb{\lambda} = (\pmb{X} + \pmb{Y})^{-1}(\pmb{x} - \pmb{y} + {}_\textrm{A}\pmb{t}_\textrm{B}).
\end{equation}

To assess the convergence, we consider the change of the translational difference between frame $\textrm{A}$ and $\textrm{B}$ and write
\begin{equation}\label{eq:a409}
	\Delta\pmb{t} = -\pmb{\theta}_{\textrm{t}a} + {}_{\textrm{A}}\pmb{R}_\textrm{B}\,\pmb{\theta}_{\textrm{t}b}.
\end{equation}
Introducing the variation vectors from (\ref{eq:a404}) and (\ref{eq:a405}) then leads to the relation
\begin{equation}\label{eq:a410}
	\Delta\pmb{t} = -(\pmb{X} + \pmb{Y})\pmb{\lambda} + \pmb{x} - \pmb{y}.
\end{equation}
Finally, with the definition of  $\pmb{\lambda}$ from (\ref{eq:a408}), we are able to write the following result
\begin{equation}\label{eq:a411}
	\Delta\pmb{t} = -{}_\textrm{A}\pmb{t}_\textrm{B}.
\end{equation}
Similar to the rotational case, this demonstrates that a single iteration of the Newton method is enough to overcome the translational error ${}_\textrm{A}\pmb{t}_\textrm{B}$ and obtain an exact kinematic solution.

\section{Adjoint Equivalence}\label{sec:a2}
In the following, we demonstrate that if the constraint coordinate frames $\textrm{A}$ and $\textrm{B}$ are equal, the first-order derivatives of the constraint equation in (\ref{eq:k23}) and (\ref{eq:k24}) reduce to an adjoint representation.
Given that in such a case the variation matrix $\pmb{C}$ is equal to the identity matrix, only the partial derivatives of the translational difference ${}_{\textrm{A}}\pmb{t}_{\textrm{B}}$ with respect to the rotational variation vectors $\pmb{\theta}_{\textrm{r}a}$ and $\pmb{\theta}_{\textrm{r}b}$ differ from the adjoint representation defined in (\ref{eq:k02}).
However, we can show that
\begin{align}\label{eq:a200}
	{}_\textrm{A}\pmb{R}_{\textrm{M}_a}\,[{}_{\textrm{M}_a}\pmb{t}_\textrm{B}]_\times &= {}_\textrm{A}\pmb{R}_{\textrm{M}_a}\,[{}_{\textrm{M}_a}\pmb{t}_\textrm{A}]_\times\,{}_\textrm{A}\pmb{R}_{\textrm{M}_a}^{-1}\,{}_\textrm{A}\pmb{R}_{\textrm{M}_a}\\[5pt]\label{eq:a201}
	&= [{}_\textrm{A}\pmb{R}_{\textrm{M}_a}\,{}_{\textrm{M}_a}\pmb{t}_\textrm{A}]_\times\,{}_\textrm{A}\pmb{R}_{\textrm{M}_a}\\[5pt]\label{eq:a202}
	&= -[{}_\textrm{A}\pmb{t}_{\textrm{M}_a}]_\times\,{}_\textrm{A}\pmb{R}_{\textrm{M}_a},
\end{align}
and similarly that 
\begin{equation}\label{eq:a203}
	-{}_\textrm{A}\pmb{R}_{\textrm{M}_b}\,[{}_{\textrm{M}_b}\pmb{t}_\textrm{B}]_\times = [{}_\textrm{A}\pmb{t}_{\textrm{M}_b}]_\times\,{}_\textrm{A}\pmb{R}_{\textrm{M}_b}.
\end{equation}
This demonstrates that if $\textrm{A}$ and $\textrm{B}$ are equal, the derivatives in (\ref{eq:k23}) and (\ref{eq:k24}) turn into $-\Ad({}_\textrm{A}\pmb{T}_{\textrm{M}_a})$ and $\Ad({}_\textrm{A}\pmb{T}_{\textrm{M}_b})$.

\section{Derivatives of Orthogonality Constraints}\label{sec:a5}
In the following, we calculate the first-order derivatives of the rotational orthogonality constraint equation.
They are required for the calculation of the constraint Jacobian in (\ref{eq:k22}).
Starting from (\ref{eq:e20}), the constraint equation is defined as
\begin{equation}\label{eq:a500}
	\begin{split}
		b_{\textrm{r}ijab}(\pmb{\theta}_a, \pmb{\theta}_b) = {}_\textrm{A}\pmb{e}_i^\top 
		&\exp\big([-{}_{\textrm{A}}\pmb{R}_{\textrm{M}_a}\,\pmb{\theta}_{\textrm{r}a}]_\times\big)\\
		&\exp\big([{}_{\textrm{A}}\pmb{R}_{\textrm{M}_b}\,\pmb{\theta}_{\textrm{r}b}]_\times\big)\\
		&{}_\textrm{A}\pmb{R}_\textrm{B}\,{}_\textrm{B}\pmb{e}_j,
	\end{split}
\end{equation}
with the tuple $(i,j) \in \{(\textrm{x}, \textrm{y}), (\textrm{y}, \textrm{z}), (\textrm{z}, \textrm{x})\}$.
Similar to (\ref{eq:31a}) and (\ref{eq:31b}), we first linearize using the Taylor series expansion
$\exp([\pmb{\theta}]_\times) \approx \pmb{I} + [\pmb{\theta}]_\times$ and then use $[\pmb{x}]_\times\pmb{y} = -[\pmb{y}]_\times\pmb{x}$.
With this, the following first-order derivatives can be calculated
\begin{align}\label{eq:a501}
		\frac{\partial{}b_{\textrm{r}ijab}}{\partial\pmb{\theta}_{\textrm{r}a}}\bigg\vert_{\pmb{\theta}_\textrm{k}=\pmb{0}}
		&={}_\textrm{A}\pmb{e}_i^\top [{}_\textrm{A}\pmb{R}_\textrm{B}\,{}_\textrm{B}\pmb{e}_j]_\times\, {}_{\textrm{A}}\pmb{R}_{\textrm{M}_a},\\[5pt]\label{eq:a502}
		\frac{\partial{}b_{\textrm{r}ijab}}{\partial\pmb{\theta}_{\textrm{r}b}}\bigg\vert_{\pmb{\theta}_\textrm{k}=\pmb{0}}
		&={}_\textrm{A}\pmb{e}_i^\top [{}_\textrm{A}\pmb{R}_\textrm{B}\,{}_\textrm{B}\pmb{e}_j]_\times\, {}_{\textrm{A}}\pmb{R}_{\textrm{M}_b}.
\end{align}
Knowing that first-order derivatives with respect to the translational variations $\pmb{\theta}_{\textrm{t}a}$ and $\pmb{\theta}_{\textrm{t}b}$ are zero, we are able to compute the constraint Jacobians $\pmb{B}_{ab}$ required in the Newton optimization.


\ifCLASSOPTIONcaptionsoff
  \newpage
\fi



\bibliographystyle{IEEEtran}
\bibliography{IEEEabrv,content/literature}
%



%


\begin{IEEEbiography}[{\includegraphics[width=1in,height=1.25in,clip,keepaspectratio]{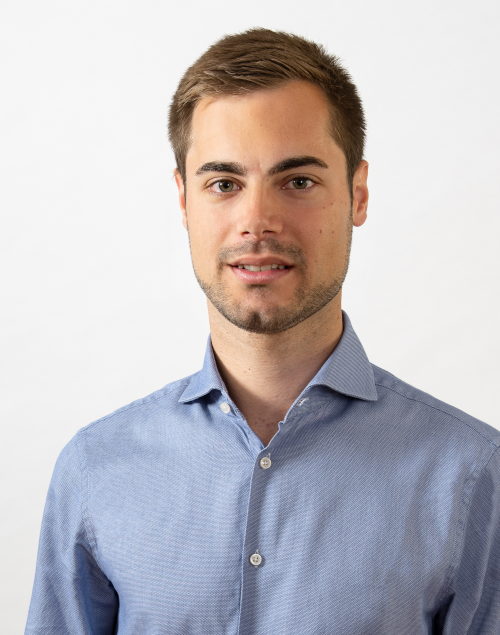}}]{Manuel Stoiber}
received a B.Sc. and M.Sc. degree from the Technical University of Munich and spent time at the Queensland University of Technology and ETH Zurich.
Currently, he is a PhD student at the Technical University of Munich and a research scientist at the German Aerospace Center, where he is part of the Perception and Cognition department in the Institute of Robotics and Mechatronics.
In 2020, he received the Best Paper Saburo Tsuji Award from the Asian Conference on Computer Vision.
His research interests include computer vision techniques and probabilistic methods for real-world robotic applications.
\end{IEEEbiography}

\begin{IEEEbiography}[{\includegraphics[width=1in,height=1.25in,clip,keepaspectratio]{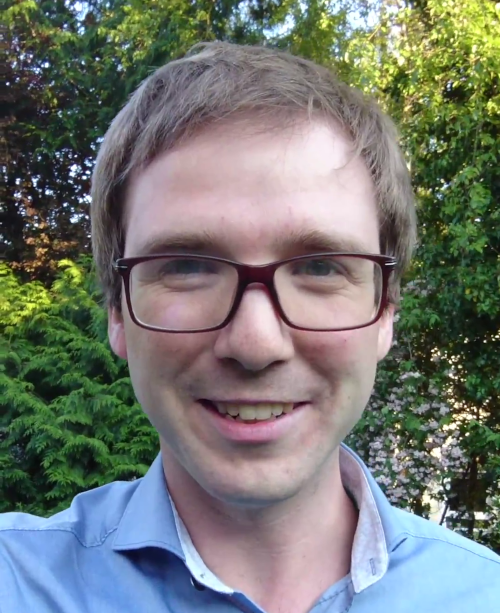}}]{Martin Sundermeyer} is a PhD student at the Technical University of Munich and a research scientist at the German Aerospace Center, where he is part of the Perception and Cognition department in the Institute of Robotics and Mechatronics. He also  performed research at the University of Tokyo and NVIDIA, received the Best Paper Award at ECCV 2018, is a core developer of BlenderProc and organizer of 6D pose estimation workshops and challenges. His research interests include computer vision methods for spatial robotic tasks with known and unknown objects, self-supervised learning from demonstrations and sim2real transfer.
\end{IEEEbiography}

\vfill

\begin{IEEEbiography}[{\includegraphics[width=1in,height=1.25in,clip,keepaspectratio]{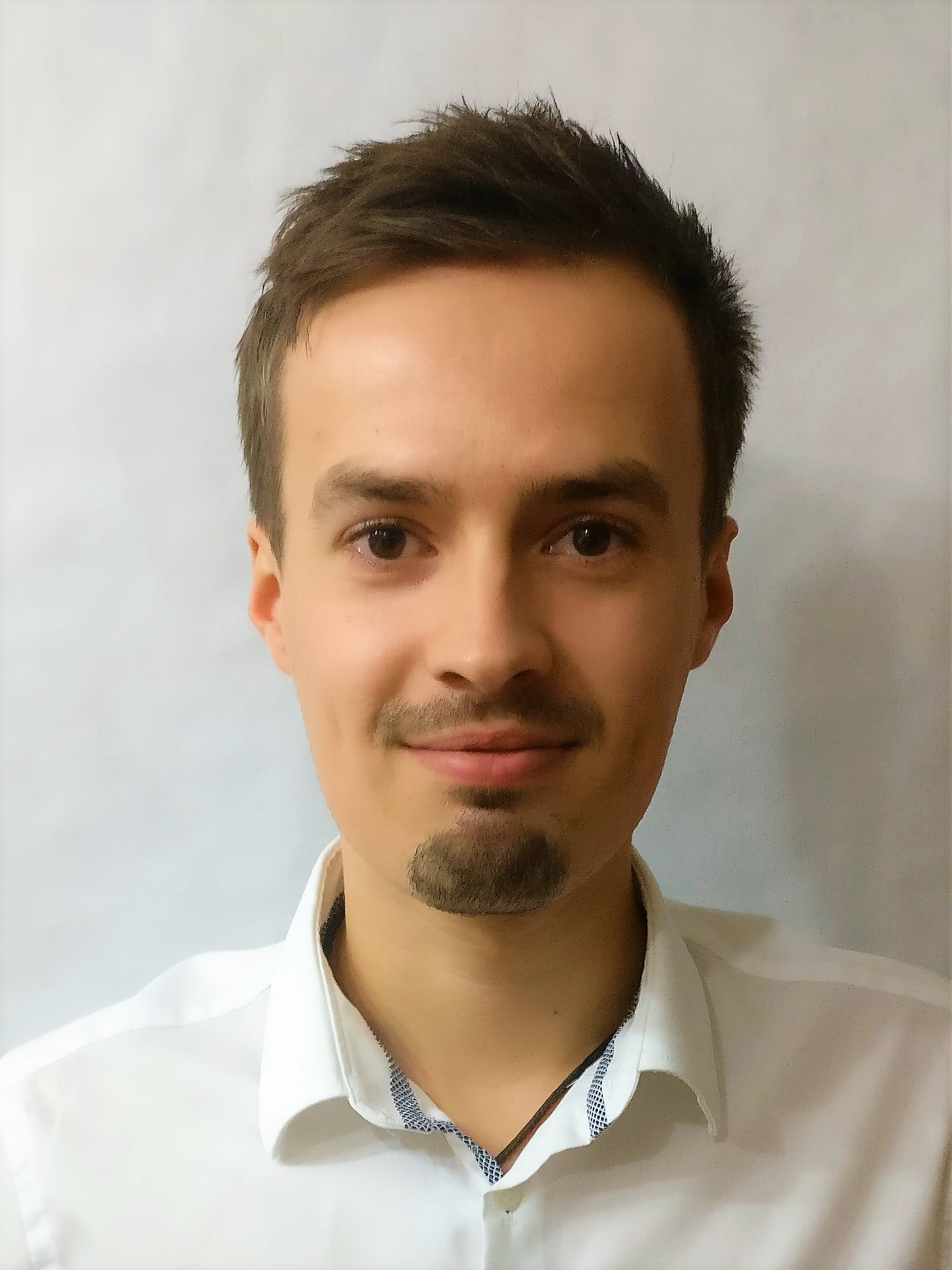}}]{Wout Boerdijk} is a PhD student at the Technical University of Munich and a research scientist at the German Aerospace Center, where he is part of the Perception and Cognition department in the Institute of Robotics and Mechatronics. His research interests include computer vision methods for learning of and interacting with objects.
\end{IEEEbiography}

\begin{IEEEbiography}[{\includegraphics[width=1in,height=1.25in,clip,keepaspectratio]{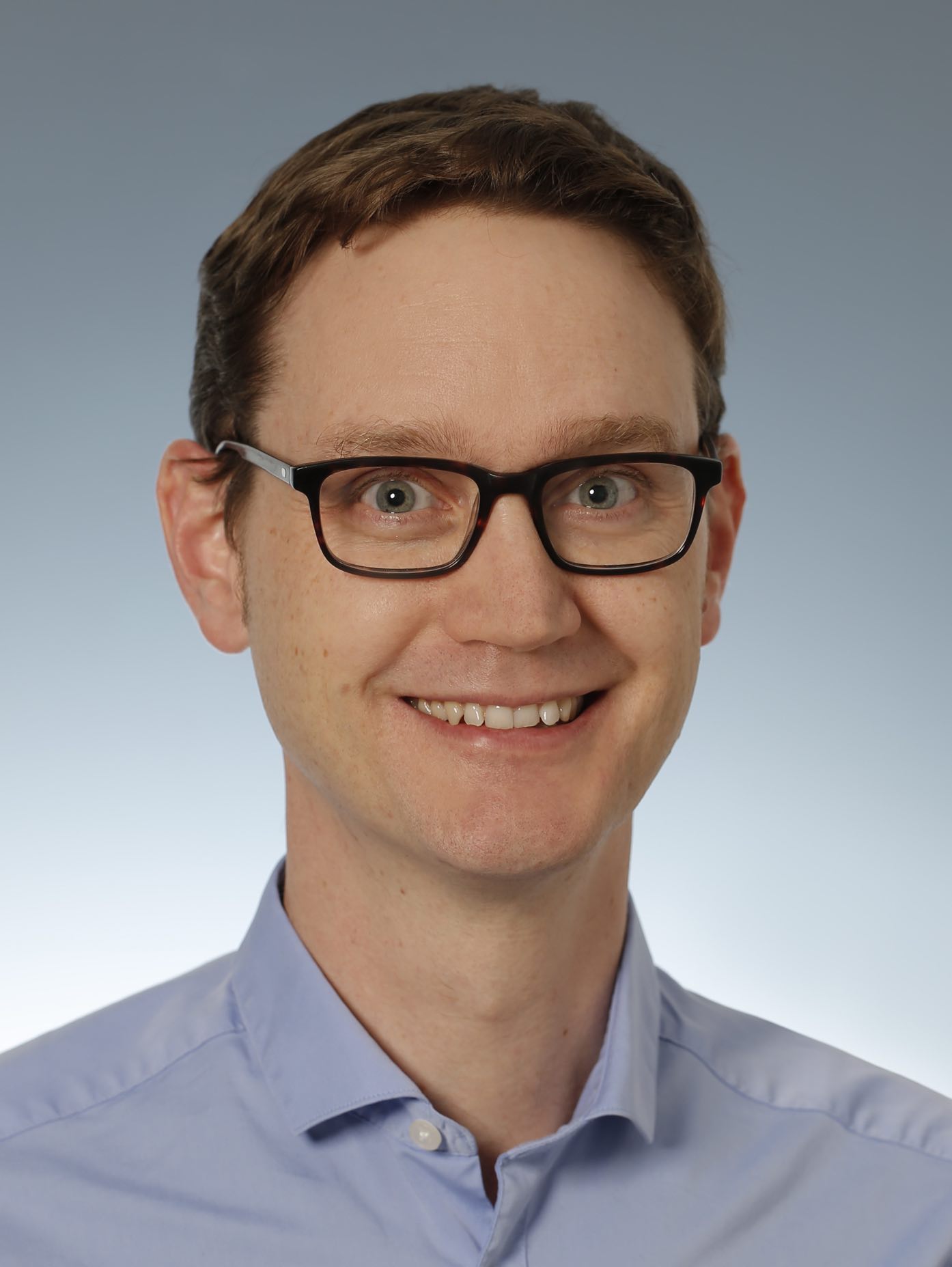}}]{Rudolph
    Triebel} leads the department of Perception and Cognition at the
  DLR Institute of Robotics and Mechatronics. He received his PhD in
  computer science in 2007 from the University of Freiburg, Germany
  and the habilitation in 2015 from the Technical University of Munich. Before working at
  DLR, he was a postdoctoral researcher at
  ETH Zurich and at the University of Oxford, UK. From 2013 to 2021,
  he was also appointed as a lecturer in computer science at TU
  Munich. Since the beginning of 2022, he is appointed as a guest
  professor in the TUM School of Engineering and Design. 
\end{IEEEbiography}


\vfill


\end{document}